\documentclass[twocolumn, switch]{article} 

\usepackage{preprint}
\usepackage{color}
\usepackage{amsmath}
\usepackage{amssymb}
\usepackage{subcaption}
\usepackage[labelfont=bf]{caption}
\usepackage{booktabs, array, multirow}
\usepackage{algorithm}
\usepackage[noend]{algpseudocode}
\usepackage{algorithmicx}
\usepackage{listings}
\usepackage{fancyvrb}
\fvset{fontsize=\small}
\usepackage
[acronym,smallcaps,nowarn,section,nogroupskip,nonumberlist]{glossaries}
\glsdisablehyper{}
\usepackage{threeparttable}
\usepackage{graphicx}
\usepackage{url}

\usepackage[numbers,square]{natbib}
\usepackage[utf8]{inputenc}	
\usepackage[T1]{fontenc}	
\usepackage{xcolor}		
\usepackage[colorlinks]{hyperref}	
\usepackage{booktabs} 		
\usepackage{nicefrac}		
\usepackage{microtype}		
\usepackage{lineno}		
\usepackage{float}			
\usepackage{appendix}

\usepackage{newfloat}
\DeclareFloatingEnvironment[name={Supplementary Figure}]{suppfigure}
\usepackage{sidecap}
\sidecaptionvpos{figure}{c}

\usepackage{titlesec}
\titlespacing\section{0pt}{12pt plus 2pt minus 2pt}{1pt plus 1pt minus 1pt}
\titlespacing\subsection{0pt}{10pt plus 2pt minus 2pt}{1pt plus 1pt minus 1pt}
\titlespacing\subsubsection{0pt}{8pt plus 2pt minus 2pt}{1pt plus 1pt minus 1pt}

\usepackage{graphics}
\usepackage{hyperref}
\usepackage{color}
\usepackage{multirow}
\usepackage{hhline}

\usepackage[nameinlink]{cleveref}
\title{\huge Deep Causal Reasoning for Recommendations}

\usepackage{authblk}

\author{Yaochen Zhu}
\author{Jing Yi}
\author{Jiayi Xie}
\author{Zhenzhong Chen\thanks{\tt{zzchen@ieee.org}}}
\affil{Wuhan University}

\begin{document}

\twocolumn[ 
  \begin{@twocolumnfalse} 
  
\maketitle

\begin{abstract}

Traditional recommender systems aim to estimate a user's rating to an item based on observed ratings from the population. As with all observational studies, hidden confounders, which are factors that affect both item exposures and user ratings, lead to a systematic bias in the estimation. Consequently, a new trend in recommender system research is to negate the influence of confounders from a causal perspective. Observing that confounders in recommendations are usually shared among items and are therefore multi-cause confounders, we model the recommendation as a multi-cause multi-outcome (MCMO) inference problem. Specifically, to remedy confounding bias, we estimate user-specific latent variables that render the item exposures independent Bernoulli trials. The generative distribution is parameterized by a DNN with factorized logistic likelihood and the intractable posteriors are estimated by variational inference. Controlling these factors as substitute confounders, under mild assumptions, can eliminate the bias incurred by multi-cause confounders. Furthermore, we show that MCMO modeling may lead to high variance due to scarce observations associated with the high-dimensional causal space. Fortunately, we theoretically demonstrate that introducing user features as pre-treatment variables can substantially improve sample efficiency and alleviate overfitting. Empirical studies on simulated and  real-world datasets show that the proposed deep causal recommender shows more robustness to unobserved confounders than state-of-the-art causal recommenders. Codes and datasets are released at \url{https://github.com/yaochenzhu/deep-deconf}.

\end{abstract}

\vspace{0.4cm}

  \end{@twocolumnfalse} 
] 

\section{Introduction}\label{sec:introduction}
\begin{figure}
\centering
\includegraphics[width=0.435\textwidth]{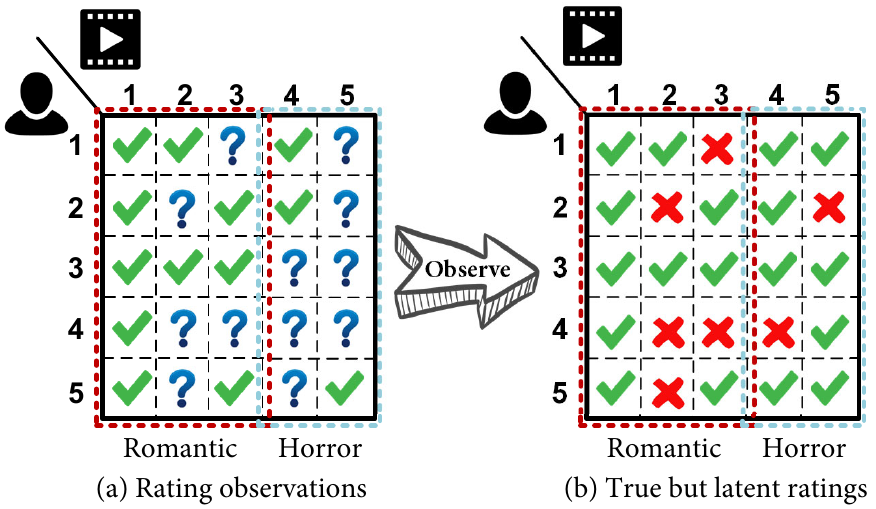}
\caption{An intuitive example of confounding bias in movie recommendation. Although users like to watch horror movies, they are seldomly exposed to the users, which leads to their under-representation in the data. Consequently, traditional recommender systems that ignores confounders may be unwilling to recommend horror movies. } 
\label{fig:teaser}
\end{figure}

\noindent Estimating users' preference based on their past behaviors is important in recommendations. Collaborative filtering, which aims at estimating one user's rating to items based on observed ratings from the population, has been widely applied in modern recommender systems. However, since a user's rating to an item is generally not independent of the item's exposure to the user, the collected rating data are unavoidably biased. Consider movie recommendation as an example. Since the genre of a movie affects both the likelihood of its exposure and user's rating to it, a spurious dependence is created between them, which makes movies in the minority genres systematically under-represented by the collected data (Fig. (\ref{fig:teaser})). This is a confounding phenomenon, and the movie genre is one of the confounders. Ignoring such confounders could lead to systematic confounding bias that degenerates the recommendation quality for traditional recommender systems \cite{chang2021bundle}. {\let\thefootnote\relax\footnote{Corresponding author: Zhenzhong Chen, E-mail: zzchen@ieee.org}}

If we could observe a user's rating to an item with and without the item's exposure to the user, confounding bias can be eliminated even in the presence of unobserved confounders \cite{rubin1990formal}. However, since whether or not an item is exposed to a user is pre-determined after the collection of the historical data, the user's rating associated with one exposure status of the item must be unobserved. Therefore, eliminating the confounding bias demands us to answer a counterfactual problem, \textit{i.e.,} what a user's rating would be if a previously unexposed item is made exposed (recommended) to her. This falls under the scope of causal inference, which aims to unbiasedly estimate unobserved exposure effects for a unit from observed outcomes from the population. In this article, the Rubin causal model \cite{rubin1990formal} is adopted as the causal inference framework, where the exposure of an item is likened to treatment in a clinical trial and the rating is likened to a potential outcome associated with an exposure. Then, recommendation can be framed as estimating the unit-level "treatment effects" for a user from observed ratings from  all users (\textit{i.e.,} the population) \cite{schnabel2016recommendations}.

To address such a counterfactual inference problem, classical causal inference demands us to find, measure, and adjust the influence of all the confounders. However, since any attribute that is shared among items (such as the genre of movies) could serve as a potential confounder, they are infeasible to enumerate. Alas, whether or not we have indeed exhausted all the confounders is not testable from experiments \cite{rubin1990formal}.

Observing that the item pools of modern recommender systems are large, we assume that single-cause confounder, which is an attribute that exclusively affects the exposure and rating of one item, is negligible. Therefore, the problem can be simplified because multi-cause confounders, \textit{i.e.,} confounders shared among items, can be properly handled by controlling and estimating substitute confounders inferred from exposures via latent factor models \cite{wang2019blessings}. This has been explored by Wang \textit{et al.} \cite{wang2020causal} by proposing the Deconfounded Recommender (Deconf-MF). However, they mainly focused on using shallow methods such as Poisson matrix factorization to infer the substitute confounders, where a closed-form approximate inference solution can be deduced. This is limited, since they may fail to capture the complex item co-exposure relationship caused by confounders. Even if the exposure model is correctly specified, the potential rating prediction model of Deconf-MF degenerates into a single-cause case where the item co-recommendation effect (the interference effect of recommending several items simultaneously) is ignored. Utilizing deep neural networks (DNNs) to model both the item exposure and user ratings, which is demonstrated to be superior in classical item co-purchase prediction tasks \cite{liang2018variational}, remains under-explored in causal recommender systems due to intractable posteriors. In addition, the multiple causes induce an exponentially large causal space, which makes the observed ratings associated with one specific exposure vector in the population scarce. This could lead to a large variance in the estimated causal effects of seldomly observed item exposures. The variance, for a DNN-based causal model, appears as its tendency to overfit when the number of user is limited.

To address the above challenges, we propose a deep deconfounded recommender (Deep-Deconf). Deep-Deconf frames the recommendation as a multi-cause multi-outcome (MCMO) inference problem, where the item exposures and user ratings are regarded as multi-cause treatments and potential outcomes, respectively. Under this modeling, item co-exposures are used to eliminate the confounding bias and item co-recommendation effects are considered to predict new ratings. Specifically, to eliminate the bias, based on a no single-cause confounder assumption, we infer and control user-specific latent variables as substitute confounders. The generative distribution is parameterized by a DNN with factorized logistic likelihood and the intractable posteriors are approximated by variational inference. Furthermore, we demonstrate that  MCMO modeling may suffer from high variance, mainly due to the scarcity of observation in high-dimensional item causal spaces. Fortunately, we demonstrate that introducing user features as pre-treatment variables, which are factors that are independent of causes but are informative to predicting potential ratings, such as user ages, genders, locations, \textit{etc.}, can substantially improve the sample efficiency. Finally, we proposed the theory of "duality of multi-cause confounders" to explain a previously ignored phenomenon where the recommendation performance increases first and then decreases with the rise of confounding levels. We demonstrate that this is because multi-cause confounders contain useful collaborative information, but greedily exploiting them could bias the model and degenerate the recommendation quality. Extensive experiments conducted on multiple simulated and real-world datasets show that Deep-Deconf is more robust to unobserved confounders than state-of-the-art causal recommenders. The main contribution is summarized as follows:

\begin{itemize}
    \item We present Deep-Deconf, a deep causality-aware recommendation algorithm based on Rubin's causal framework. Through controlling a special user latent factor and informative user features as pre-treatment variables, Deep-Deconf leads to confounding-bias-robust recommendation with low variance.
    \item Theoretical analysis demonstrates the global and local network Jacobians of Deep-Deconf can be associated with average recommendation effects for all and a subsection of the population, where the explainability of Deep-Deconf can be guaranteed.
    \item We demonstrate the large variance associated with the multi-cause causal inference modeling paradigm. As a solution, we theoretically and empirically show that controlling user features as pre-treatment variables can substantially lower the variance.

    \item Experiments conducted on simulated and real-world datasets demonstrate that Deep-Deconf is robust to confounding bias. Moreover, we propose the theory of "duality of multi-cause confounders" to explain the non-monotone change of recommendation performance with the increase of confounding levels. 
\end{itemize}

The remainder of this paper is as follows. Section \ref{seq:related} surveys related work regarding causal recommenders and deep causal inference techniques. Section \ref{seq:meth_form} provides formal formulation of the causal recommendation problem. Section \ref{seq:meth} expounds upon the proposed Deep-Deconf model in details. Section \ref{seq:emp} demonstrates the extensive experiments conducted on simulated and real-world datasets as well as the theory of duality of multi-cause confounders. Finally, section \ref{seq:conc} concludes our paper.

\section{Related Work}
\label{seq:related}

\subsection{Causality-based Recommendations} \noindent Recently, researchers have been aware that recommendation is an intervention analogous to treatment in clinical trials \cite{schnabel2016recommendations}. Since randomized experiments are clearly infeasible for recommendations, confounders, which are factors that affect both the item exposure and user ratings, pervasively exist and bias the collected data \cite{steck2013evaluation}. Fortunately, causal inference can be utilized to eliminate the confounding bias and uncover the true causal relationships between two variables \cite{ schnabel2016recommendations}. Existing causality-based recommender systems can be classified into three main categories: propensity score reweighting (PSW)-based methods, substitute confounder controlling-based methods, and graph adjustment-based methods. In addition, methods from each category can also use two paradigms as the fundamental causal framework: Rubin's potential outcome causal framework (RCF) \cite{imbens2015causal}, and Pearl's structural causal graph (SCG) framework \cite{pearl2016causal}. 

The concept of PSW originated from RCF \cite{imbens2015causal}, the core of which is that outcomes (ratings) under different treatment status, i.e., potential outcomes, follow different distributions, where one of which is unavoidably unobservable. Therefore, to calculate the treatment effects, i.e., the difference between potential outcome under treatment and under no treatment, randomized experiments should be conducted to select units in treatment and non-treatment groups such that units in two groups are comparable to each other. However, random recommendations are clearly infeasible in modern online systems. Consequently, unobserved confounders pervasively exist and cause discrepancies between the two groups in the collected data. PSW aims to reweight users in the treatment group by the chances that they receive the treatment (i.e., the propensity scores), such that they can be viewed as random samples from the population \cite{ salakhutdinov2010coL, joachims2017unbiased, bonner2018causal}.  Linear regression \cite{schnabel2016recommendations} and variational auto-encoders (VAEs) \cite{zou2020counterfactual} have been used to estimate the propensity scores from historical ratings and user features. However, one problem of PSW is that the estimated propensity scores can be extremely small when the causal space is high-dimensional, which leads to large sample weights that make the training dynamics unstable. Moreover, the unbiasedness of PSW relies on a correctly-specified propensity score estimation model. However, the model design depends heavily on the expertise of the researcher and is therefore untestable from experiments.  

Another class of causal recommenders based on the RCF aims to find, measure, and control all the unobserved confounders. However, due to the clear infeasibility of the objective, substitute confounders are inferred from item exposures and are controlled as surrogates to the true confounders. One exemplar work is Deconf-MF \cite{wang2020causal}, where the item co-recommendations are viewed as a bundled treatment where latent factors are estimated to render item exposures conditionally independent. Under mild assumptions, controlling such latent factors as substitute confounders can be proved to eliminate the confounding bias. However, both the confounder inference model and recommendation model used in \cite{wang2020causal} are based on shallow matrix factorization, which may have insufficient modeling ability for large-scale modern recommendation tasks. Utilizing deep neural networks (DNNs) to model both the item exposure and user ratings remains under-explored due to intractable posteriors. Faced with this challenge, we generalize Deconf-MF to a DNN-based framework where the non-linear influence of confounders to the ratings can be captured and remedied, and this is one of the core contributions of Deep-Deconf. Concurrent with our work, \cite{ma2021multi} proposed the DIRECT algorithm to extend Deconf-MF, which focuses on learning disentangled representations of the treatment with VAE for better explanation. Compared to DIRECT, Deep-Deconf mainly tackles network weights interpretability, variance reduction of MCMO modeling, and duality of multi-cause confounders, which are three new independent research questions worthy of in-depth investigations.

Relying on Pearl's SCG framework, graph adjustment-based methods construct a priori a graph that depicts the causal relationships among relevant variables. Compared to RCF, relationships among user/item features, interaction histories, item exposures, and user ratings can be clearly expressed as nodes in the causal graph. To eliminate confounding effects, backdoor adjustment is generally applied on the graph, where pre-treatment factors that affect the treatment assignment (i.e., item exposure) are eliminated from the graph such that items can be viewed as exposed in a random manner. Generally, the established SCG varies drastically among different papers, both in variables included for consideration and the assumed links among them. Moreover, there is also no consensus in approximate inference strategies to solve the graph \cite{zhang2021causally,tan2021counterfactual,Zhang2021causal,zheng2021disentangling}. For example, \cite{Zhang2021causal} models the item popularity as the confounder and eliminates its influence to reduce the popularity bias. \cite{xu2021causal} explicitly models the influence of users' historical rating on items' exposure, where the influence is then removed via do-calculus. The superiority of SCG or RCF has been a controversial topic that is under constant debate among the researchers \cite{gelman2009resolving}. The reason why we build Deep-Deconf upon the substitute confounder-based method instead of Pearl's SCG is that, since the establishment of SCG is unavoidably subjective, whether it exhaustively enumerates all confounders or correctly specifies their relationships is untestable. In contrast, in substitute confounder-based methods the causal model can be agnostic of the specific form and relationships among the confounders, which eliminates the strong substantive assumptions of SCGs.

\subsection{Deep Learning for Causal Inference} \noindent Recent years have witnessed an upsurge in interest in utilizing DNNs for traditional causal inference problems \cite{xu2019scalable, pawlowski2020deep, luo2020causal, scholkopf2021toward}. Among them, the most relevant methods to Deep-Deconf are \cite{louizos2017causal} and \cite{rakesh2018linked}, which explored the deep latent-variable model for single cause inference tasks, \textit{e.g.,} twin weights and job training, based on RCF. They mainly focused on modeling the joint distribution of treatment, hidden confounders, and potential outcomes via deep generation networks and approximating the intractable posteriors of hidden confounders through variational encoders. However, this strategy is not directly applicable to recommender systems, because when multiple causes exist, it requires a sub-encoder and a sub-decoder for each treatment configuration. But since the number of treatments for recommendations is exponential to the number of items, extrapolating all the missing potential outcomes requires intractable numbers of inference and generation networks. The proposed Deep-Deconf circumvents this issue by two-stage modeling, where the exposure model first fits the joint distribution of substitute confounder and item exposures, and the outcome model then fits the rating distribution conditional on the substitute confounder, item exposures, and user features. This leads to a parameter-efficient solution to the DNN-based multiple cause inference problems.

\section{Problem Formulation}
\label{seq:meth_form}
\noindent Suppose a system with $U$ users and $I$ items. The observational data comprises the rating matrix $\mathbf{R} \in \mathbb{R}^{U \times I}$ and the user features $\mathbf{X} \in \mathbb{R}^{U \times S}$ where each row $\mathbf{r}^{T}_{u}$, $\mathbf{x}^{T}_{u}$ are the rating and feature vector for user $u$.\footnote{\small{Symbol system: We use boldface capital symbols $\mathbf{R}$ for matrices, boldface lower case symbols $\mathbf{r}_{u}$ for vectors, non-boldface lowercase symbols $r_{ui}$ for scalars, non-boldface capital symbols $R_{u}$ for random variables ($R_{u}$ may be scalar, vector, or matrix based on context). Exceptions such as $U$ and $I$ can be easily identified from the context.}} Based on the Rubin causal model \cite{imbens2015causal}, the received treatment is represented by the exposure matrix $\mathbf{A}$ where $a_{ui}$ denotes whether the item $i$ has been exposed to the user $u$ when rating $r_{ui}$ is provided. We denote the potential outcome random variable associated with an exposure $A_{u}$ for the user $u$ as $R_{u}(A_{u})$. For each user, we only observe the value of $R_{u}(A_{u}=\mathbf{a}_{u})$, \textit{i.e.,} $\mathbf{r}_{u}(\mathbf{a}_{u})$. The main quantity of interest is the ratings user $u$ would provide if $K$ extra items are recommended. Therefore, recommendation under causal reasoning is a counterfactual inference problem. To address this problem, we assume that stable unit treatment value assumption (SUTVA) holds where the ratings of user $u$ is independent of items' exposure to user $v$, \textit{i.e.}, $R_{u} \perp A_{v} \mid A_{u}$. This excludes from consideration the interference among users. The randomness of $R_{u}$ and $A_{u}$ sources from the fact that the user $u$ is an arbitrary user sampled from a potentially very large super-population. Therefore, unless specified otherwise, the expectations are taken w.r.t. this population for the rest of the article. The purpose of this article is to estimate the expected causal effects of an exposure $A_{u}$ to a user $u$, $\mathbb{E}[R_{u}(A_{u})]$, from the population so that unbiased recommendations can be made accordingly based on estimated ratings with low variance.

\section{The Deep Deconfounded Recommender}
\label{seq:meth}

\subsection{Debias via Factorized Variational Auto-encoder}
\label{seq:meth_debias}

\begin{figure}
\centering
\includegraphics[width=0.45\textwidth]{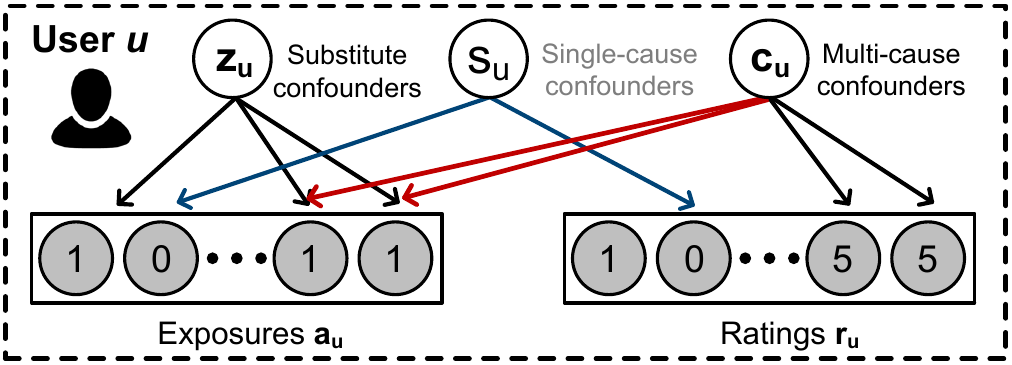}
\caption{The probabilistic graphical model (PGM) for different kinds of confounders in recommender systems. The intuition of our model is to find the substitute confounders $\mathbf{z}_{u}$ that render the item exposures $\mathbf{a}_{u}$ independent Bernoulli trials. Then, conditional on $\mathbf{z}_{u}$, multi-cause confounders cannot exist. The proof is by contradiction (see the red lines).} 
\label{fig:pgm}
\end{figure}

\noindent  
Non-causality-based recommender systems directly model the conditional distribution $p(R_{ui}\mid A_{ui})$ via matrix factorization or DNNs and fit it on observed ratings to calculate $\mathbb{E}[R_{ui}(A_{ui}=a_{ui})]$, relying on the assumption that $\mathbb{E}[R_{ui}(A_{ui}=a_{ui})] = \mathbb{E}[R_{ui}(A_{ui}) \mid A_{u} = \mathbf{a}_{u}]$, \textit{i.e.,} the exposure of the items does not depend on user $u$'s rating toward them. The estimation is unbiased if and only if there exist no variables that simultaneously affect $R_{ui}$ and $A_{u}$, \textit{i.e.,} no unobserved confounders \cite{wang2019blessings}. However, unobserved confounders pervasively exists in collected ratings, which leads to systematic bias (\textit{e.g.,} amplified exposure bias due to similar bias in similar users) for naive methods \cite{steck2010training}.

To eliminate confounding bias, classical causal inference techniques demand us to find, measure and control all confounders $C_{ui}$ and calculate $\mathbb{E}[R_{ui}(A_{ui})]$ from  $\mathbb{E}_{C_{ui}}\left[\mathbb{E}\left[ R_{ui}(A_{ui}) \mid C_{ui}, A_{u} \right]\right]$. However, it can guarantee unbiasedness provided that there are no uncontrolled confounders. This is known as the strong ignorability assumption, which is both infeasible and untestable from experiments \cite{holland1986statistics}.  Therefore, to circumvent exhausting and measuring all confounders $C_{ui}$, Deep-Deconf models the recommendation as a multi-cause inference problem. Instead of treating the exposure of an item $A_{ui}$ as an isolated cause, we consider the exposures of all the items to a user $A_{u}$ as a holistic treatment that could causally affect all the ratings. We first assume that the single-cause confounder $S_{u}$, which influences only the exposure of one specific item and its ratings, does not exist. The validity of the assumption can be justified by the fact that the pool of candidate items for modern recommender systems is usually large, and therefore it is unlikely that a confounder influences only one of the items. Taking genre as an example, the genre of a movie affects both its exposure and rating, and it is a universal attribute that is shared among all movies. With the assumption of no-single cause confounders, we only need to control the multi-cause confounders. This is a more amenable objective, since controlling them is equivalent to controlling latent factors $Z_{u} \in \mathbb{R}^{K}$ that render the causes conditionally independent, \textit{i.e.,}  $p(A_{u} \mid Z_{u}) = \Pi_{i}p(A_{ui} \mid Z_{u})$. A simple proof for the validity of the claim is that, if multi-cause confounders still exist after conditioning on such latent factors, the exposures cannot be conditionally independent, which renders a contradiction. An intuition can be referred to in Fig. (\ref{fig:pgm}).

To find such latent variable $Z_{u}$, we first parameterize the generative distribution $p_{\theta}(A_{u} \mid Z_{u})$ by a DNN with factorized logistic likelihood,  \textit{i.e.,} $p(\mathbf{a}_{u} \mid \mathbf{z}_{u}) = \Pi_{i=1}^{I} Bern(a_{ui} \mid [\theta(\mathbf{z}_{u})]_{i})$. Conditional on $Z_{u}=\mathbf{z}_{u}$, the exposures $a_{ui}$ for user $u$ can be viewed as generated from randomized Bernoulli trials. The factorized Bernoulli distribution satisfies the overlap assumption, \textit{i.e.,} $p_{\theta}(A_{u} \mid Z_{u}) > 0$ provided that $[\theta(\mathbf{z}_{u})]_{i} \notin \{0,1\}$, which is crucial to the identifiability of the model \cite{imai2004causal}.  We then calculate the intractable posterior $q_{\phi}(Z_{u} \mid A_{u})$ via the variational inference \cite{kingma2013auto}, where the prior for $Z_{u}$ is set to be the standard Normal $\mathcal{N}(\mathbf{0}, \mathbf{I}_{K})$. It could be noted that the assignment model resembles the Multi-VAE for recommendation with implicit feedback \cite{liang2018variational} because they both manage to reconstruct the binary exposure/click inputs. However, a characteristic that makes the assignment model fundamentally different is that it requires $A_{u}$ to factorize conditional on $Z_{u}$. This renders the multinomial likelihood (which has been demonstrated to be more suitable for recommendations in \cite{liang2018variational}) invalid in our case. The assumption of no single-cause confounder and the utilization of user latent factors as surrogate multi-cause confounders lead us to the following equality,
\begin{equation}
\label{eq:unconf}
  \mathbb{E}\left[R_{u}(A_{u})\right] = \mathbb{E}_{Z_{u}}[\mathbb{E}\left[R_{u}(A_{u}) \mid Z_{u},  A_{u} \right]].
\end{equation}
If the exposure model is accurately specified, conditional on $Z_{u}=\mathbf{z}_{u}$, the observed ratings for the users can be analyzed as were generated from a randomized experiment, and bias due to unobserved multi-cause confounders is eliminated. 

\subsection{Rating Prediction via Deep Outcome Network}
\label{seq:meth_ratepreds}

\begin{figure}
\centering
\includegraphics[width=0.42\textwidth]{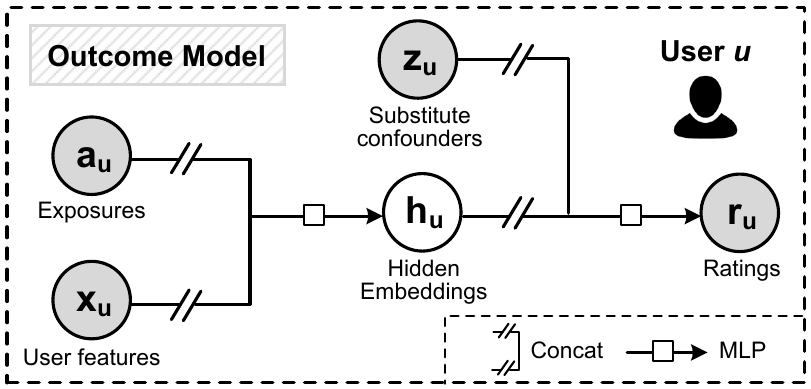}
\caption{The structure of the outcome prediction model.} 
\label{fig:out}
\end{figure}

\begin{figure}
\centering
\includegraphics[width=0.47\textwidth]{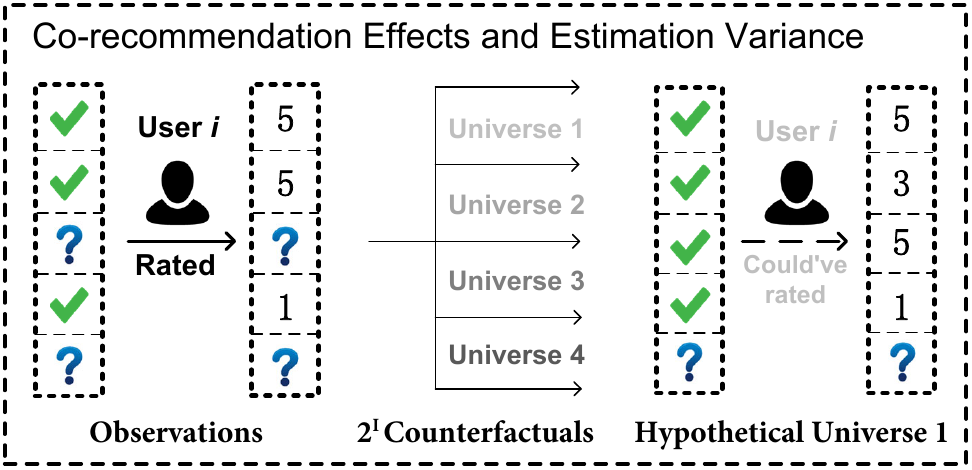}
\caption{The co-recommendation effects and large variance associated with MCMO modeling. In parallel universe 1, user $i$ had been recommended with item \#3, which increased her expectation to item \#2 and made her less satisfied with it. The large variance of MCMO is because the number of such counterfactuals is exponential to the number of items.} 
\label{fig:counter}
\end{figure}

\noindent In this section, we introduce the deep outcome network that predicts the potential ratings $R_{u}(A_{u})$. Since the outcome network no longer requires the ratings to be conditionally independent, more powerful multinomial likelihood can be put on ratings for more accurate predictions (see Fig. (\ref{fig:out}) for the structure of the deep outcome model). To gain more insights into the network, we first provide a theoretical analysis of the network weights for a special case where the network has a single layer with no activation, \textit{i.e.,} 
\begin{equation}
\label{eq:network}
    \mathbf{r}_{u}(\mathbf{a}_{u}) = \mathbf{W}^{z}\mathbf{z}_{u}+\mathbf{W}^{a}\mathbf{a}_{u} + \boldsymbol{\alpha} + \boldsymbol{\epsilon}_{u}, 
\end{equation}
where $\boldsymbol{\epsilon}_{u}$ is the residual vector for user $u$ and $\boldsymbol{\alpha}$ is the constant term. For users with inferred substitute confounder equals $\mathbf{z}_{u}$, we can calculate the the expected causal effect for the exposure of item $j$ on the rating of item $i$ as
\begin{equation}
\label{eq:ept_delta}
\begin{aligned}
    \mathbb{E} [\Delta R _ {ui} \mid \mathbf{z} _ {u}] &= \mathbb{E} [R_{ui}( \mathbf{a}+ \mathbf{e}_{j}) - R_{ui}(\mathbf{a}) \mid \mathbf{z}_{u}] \\ &= [\mathbf{W}^{a} \mathbf{e}_{j}]_{i} = w^{a}_{ij},
\end{aligned}
\end{equation}
\noindent which shows that the network weights $w^{a}_{ij}$ can be interpreted as the conditional average treatment effect (CATE) of recommending item $i$ on the rating of item $j$. Moreover, since the R.H.S. of Eq. (\ref{eq:ept_delta}),  $\mathbf{w}^{a}_{ij}$ does not contain a $\mathbf{z}_{u}$-related term, we have $\mathbb{E}_{Z_{u}}\left[\mathbb{E} \left[\Delta R_{ui} \mid Z_{u}\right]\right]= w^{a}_{ij}$, which leads to a further conclusion that $w^{a}_{ij}$ is also the average treatment effect (ATE) in the population. Note that allowing the causal effects of one item's exposure  on the rating of another item does not violate the SUTVA assumption, as the non-interference of users' exposures to each other (\textit{i.e.}, $R_{u} \perp A_{v} \mid A_{u}, \ \forall u \ne v < U $) does not exclude from consideration the interference of different items' exposures of a single user (\textit{i.e.}, $R_{ui} \perp A_{uj} \mid A_{ui}, \ \forall i \ne j < I $). On the contrary, it is benificail to model such co-recommendation effects, since exposing one movie to a user may alter her expectation to movies with a similar genre and therefore causally influences her ratings towards these movies as well. (Illustration see Fig. (\ref{fig:counter}))

\subsection{Variance Reduction via Pre-treatment Variables}
\label{seq:var_redu}

\noindent Although confounding bias can be remedied by the controlling substitute confounders, the predicted ratings may suffer from a large variance due to the following two factors: First, multiple causes lead to a more severe data missing problem. Consider a system with $I$ items. The number of counterfactuals is $2^{I}$, but only one of the outcomes is observed. Therefore, the sample efficiency is exponentially reduced compared with the classical single cause problems. Furthermore, even if the observations are sufficient, the item exposures tend to depend heavily on the user preference. This makes the exposure and non-exposure group for different treatments highly imbalanced, which further increases the estimand variance \cite{imbens2015causal}. In this section, we discuss the variance reduction technique of Deep-Deconf by introducing user features in the outcome model as pre-treatment variables, which are factors that remain uninfluenced by item exposures but are predictive to the ratings. 

To see how this works, we derive the estimand variance before and after introducing user features as pre-treatment variables. For simplification, the network weights $\mathbf{W}^{a}$ are for now reduced to diagonal, \textit{i.e.,} $\mathbf{w}^{a} = \operatorname{diag}(\mathbf{W}^{a})$, and $z_{u}$, $x_{u}$ are one-dimensional. In this scenario, the co-recommendation effects vanish and we can treat the recommendation of each item $R_{ui}$ separately (Note that the randomness of $R_{ui}$ is still solely due to the sampling of $u$ from the population since the item $i$ is fixed). To intuitively show the variance reduction, we put aside for now the powerful multinomial likelihood on ratings, and focus on the Gaussian likelihood and the simple Ordinary Least Square (OLS) optimizer. The outcome model for rating $R_{ui}$ before the introduction of user features can be specified as
\begin{equation}
\label{eq:before_pre_deep}
        r_{ui}(a_{ui}) = w^{z}z_{u}+w_{i}^{a}a_{ui} + \alpha_{i} + \epsilon_{ui}.
\end{equation}
Conditional on $Z_{u}=z_{u}$, with the overlap assumption, when the sample size $U$ is large enough, we can meaningfully calculate the following statistics from the samples,
\begin{equation}
  \bar{r}_{ui}(t) = \sum_{u: a_{ui}=t, Z_{u}=z_{u}} r_{ui} / U^{t}_{i},
\end{equation}
$\mathrm{where} \ U^{t}_{i} = \sum_{u: Z_{u}=z_{u}} \mathbb{I}(a_{ui}=t), \ t \in \{0, 1\}$ is the size of exposure and non-exposure group in the sub-population. We define CATE on the rating $r_{ui}$ as $\tau_{ui} = \mathbb{E}[r_{ui}{(1)} - r_{ui}(0) \mid z_{u}]$. Since unconfoundedness holds when conditional on $Z_{u}$, the average rating difference between exposure and non-exposure use group $\hat{\tau}_{ui} = \bar{r}_{ui}(1) - \bar{r}_{ui}(0)$ is asymptotically unbiased for $\tau_{ui}$. Furthermore, if we define $w_{i}^{ols}$ as the coefficient $w_{i}^{a}$ learned from the $U$ users drawn from the population, similar deductions as Eq. (\ref{eq:network}) can demonstrate that $w_{i}^{ols} = \hat{\tau}_{ui}$. The variance of $w_{i}^{ols}$, under the assumption of homoskedasticity, converges in probability to $\frac{\sigma_{R_{ui} \mid Z_{ui},A_{ui}}^{2}}{U \cdot p_{i} \cdot(1-p_{i})}$ when $U$ approaches infinity, where $p_{i}$ is the limit of $U^{1}_{i}/U$ and $\sigma^{2}_{R_{ui} \mid Z_{ui},A_{ui}}$ is the population conditional variance.  

After that, we consider introducing user features $\mathbf{x}_{u}$ as additional pre-treatment variables. The new model becomes
\begin{equation}
\label{eq:after_exp}
        r_{ui}(a_{ui}) = w^{z}z_{u}+w_{i}^{a}a_{ui} + w^{x}x_{u} + \alpha_{i} + \epsilon_{ui}.
\end{equation}
After introducing an additional covariate $x_{u}$ in the model, the same algebra shows that $w_{i}^{xols}$ estimated by OLS is still asymptotically unbiased for $\hat{\tau}$. But the limiting variance of the estimand becomes $\frac{\sigma_{R_{ui} \mid X_{u}, Z_{ui},A_{ui}}^{2}}{U \cdot p_{i} \cdot(1-p_{i})}$ (Proofs are in Appendix \ref{sec:proof_var_appendix}). Therefore, as long as the user features are indicative to the variation of the ratings, $\sigma_{R_{ui} \mid X_{u}, Z_{ui},A_{ui}}^{2}$ can reduce considerably compared to the marginal variance $\sigma_{R_{ui} \mid Z_{ui},A_{ui}}^{2}$, which increases the precision. 

\subsection{Theoretical Analysis in the Non-linear Case}

\noindent The previous derivations of CATE interpretation of network weights and variance reduction are mainly based on a simple linear network. However, as the main contribution of this article is proposing a deep causal model for recommendation, we generalize the analysis to the non-linear case. We begin by defining the DNN-based outcome model as
\begin{equation}
    \mathbf{r} _ u(\mathbf{a} _ u) = f_{nn}(\mathbf{a} _ u, \mathbf{z} _ u, \mathbf{x} _ u),
\end{equation}
where function $f_{nn} : \mathbb{R}^{I+K+F}  \rightarrow \mathbb{R}^{I}$ is non-linear but differentiable almost anywhere. The generalization is achieved from two aspects. First, we consider the global property of $f_{nn}$. Since the exposure $\mathbf{a} _ u$ is a binary vector, the prior of substitute confounder $\mathbf{z} _ u$ is $\mathcal{N}(\mathbf{0}, \mathbf{I}_{k})$, the user features $\mathbf{x} _ u$ are rescaled to zero mean and unit variance, we can form a global approximation of  $ f_{nn}$ with its Taylor expansion at $(\mathbf{0} ,\mathbf{0} ,\mathbf{0})$ and keep the linear term, 
\begin{equation}
 \mathbf{r} ^ {0} _ u(\mathbf{a} _ u) = f^{0}_{nn}(\mathbf{a} _ u, \mathbf{z} _ u, \mathbf{x} _ u) \approx \mathbf{W} ^{a} _ {0}\mathbf{a} _ u +  \mathbf{W} ^ {z} _ 0 \mathbf{z} _ u +  \mathbf{W} ^ {x} _ 0  \mathbf{x} _ u + \boldsymbol{\alpha}_{0},   
\end{equation}
where the coefficient matrices $\mathbf{W} ^{\{a,z,x\}} _ {0}$ are the Jacobians at $(\mathbf{0} ,\mathbf{0} ,\mathbf{0})$, and $\boldsymbol{\alpha}_{0}$ is the expected user ratings where no items are recommended. The reason to justify the approximation of $f^{0}_{nn}$ to $f_{nn}$ is that generally, $f_{nn}$ cannot be highly-nonlinear ($f_{nn}$ is generally composed of 0-2 hidden layers); otherwise, the outcome model will overfit on the negative samples which are not truly negative and fail to generalize to recommend new items \cite{liang2018variational}. After linearization, the same theoretical analysis can be applied to $f^0 _ {nn}$, where $w ^{a} _ {0, ij}$ can be interpreted as both the CATE and the ATE of the recommendation of item $i$ on the rating of item $j$.

\begin{figure}
\centering
\includegraphics[width=0.49\textwidth]{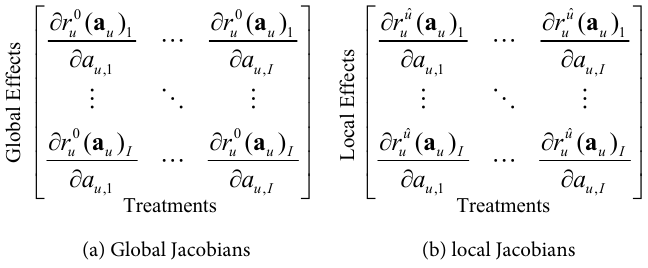}
\caption{The association between the global and local Jacobians of the outcome network and the approximate average treatment effects for all and sub-section population.} 
\label{fig:jacobian}
\end{figure}

However, if $\mathbf{a} _ {u}$ for some users is dense and deviates far away from the original point, the above approximation may be coarse and inaccurate for this sub-population. Therefore, we propose another refined generalization strategy that shows the local property of $f_{nn}$. For user $\hat{u}$ with exposures $\mathbf{a} _ {\hat{u}}$, substitute confounders $\mathbf{z} _ {\hat{u}}$, and user features $\mathbf{x} _ {\hat{u}}$, we can linearize $f_{nn}$ at the point $(\mathbf{a} _ {\hat{u}}, \mathbf{z} _ {\hat{u}}, \mathbf{x} _ {\hat{u}})$ by Taylor expansion, 
\begin{equation}
\begin{aligned}
& \mathbf{r} ^ {\hat{u}} _ u(\mathbf{a} _ u)   = f^ {\hat{u}} _ {nn} (\mathbf{a} _ {u}, \mathbf{z} _ {u}, \mathbf{x} _ {u}) \approx \mathbf{W} ^ {a} _ {\hat{u}}  (\mathbf{a} _ u - \mathbf{a} _ {\hat{u}})+ \mathbf{W} ^ {z} _ {\hat{u}} (\mathbf{z} _ {u} - \mathbf{z} _ {\hat{u}}) \\  & +  \mathbf{W} ^ {x} _ {\hat{u}}  (\mathbf{x} _ {u}-\mathbf{x} _ {\hat{u}}) +  \mathbf{r} _ {\hat{u}} = \mathbf{W} ^{a} _ {\hat{u}}\mathbf{a} _ {u} +  \mathbf{W} ^{z}_ {\hat{u}} \mathbf{z} _ {u} +  \mathbf{W} ^ {x} _ {\hat{u}}  \mathbf{x} _ {u} + \boldsymbol{\alpha} _ {\hat{u}}, 
\end{aligned}
\end{equation}
where $\mathbf{W} ^{a,z,x}_ {\hat{u}}$ are the network Jacobians at $(\mathbf{a} _ {\hat{u}}, \mathbf{z} _ {\hat{u}}, \mathbf{x} _ {\hat{u}})$, and $\boldsymbol{\alpha} _ {\hat{u}}$ is the expected ratings for user $\hat{u}$ where no items are recommended. With the local linearization of $f_{nn}$, similar analysis can be applied to $f ^ {\hat{u}} _ {nn}$. Note that the trade off is that $w ^{a}_ {\hat{u}, ij}$ is no longer the ATE of recommending item $i$ on the rating of item $j$ for the entire population, but is only the CATE for users who are similar with $\hat{u}$ (which is measured by $\mathbf{z} _ {u}$). Therefore, this strategy establishes a corresponding relationship between local Jacobians of $f_{nn}$ and CATE for a sub-population. The details are in Fig. (\ref{fig:jacobian})

\begin{algorithm}[t]
\caption{The Deep Deconfounded Recommender.}
\label{alg:all}
\begin{algorithmic}[1]
\State {\bfseries Input:}  $\mathcal{D} = \{(\mathbf{a}_{u}, \mathbf{x}_{u},  \mathbf{r}_{u}(\mathbf{a}_{u})\}$, a dataset of exposures, user features, and collected ratings, where $r_{u} \in \{1,2,3,4,5\}^{I}, \mathbf{x}_{u} \in \mathbb{R}^{S}, \mathbf{a}_{u} \in \{0,1\}^{I}, u \in \{1, 2, \ldots, U\} $.
\State \# \textit{Definition of the Exposure Network}
\Function{ExposureNet}{$\{\mathbf{a}_{u}\}$}
\For{$e=1,2,\ldots,N_{epochs}$}
\For{$u=1,2,\ldots,U_{train}$}
\State $\hat{\mathbf{z}}_{u} \sim \mathcal{N}\left(f^{\mu}_{enc}(\mathbf{a}_{u}), f^{\sigma^{2}}_{enc}(\mathbf{a}_{u})\right)$
\State $\mathbf{a}^{rec}_{u} \sim $ \Call{factorizedLogistic}{$f_{dec}(\hat{\mathbf{z}}_{u}$)}
\EndFor
\State \# \textit{Definition of the Outcome Network.}
\State $L_{train} \gets \sum_{u=1}^{U_{train}} \left(\ln p(\mathbf{a}^{rec}_{u}) - \beta \cdot \operatorname{KL}\right)$
\State Update weights $\boldsymbol{\theta}$, $\boldsymbol{\phi}$ to maximize $L_{train}$
\EndFor
\State {\bfseries Until} $L_{val} \gets \sum_{u=U_{train}}^{U_{train}+U_{val}} \ln p(\mathbf{a}^{rec}_{u})$ decreases.
\For{$u=1,2,\ldots,U,$} $\mathbf{z}_{u} \gets f^{\mu}_{enc}(\mathbf{a}_{u})$
\EndFor
\State \Return $\{\mathbf{z}_{u}\}_{u=1}^U$ as the substitute confounders.
\EndFunction
\State \# \textit{Definition of the outcome network}
\Function{OutNet}{$\{\mathbf{r}_{u}\}$, $\{\mathbf{z}_{u}\}$, $\{\mathbf{x}_{u}\}$, $\{\mathbf{a}_{u}\}$ (opt.)}
\For{$e=1,2,\ldots,N_{epochs}$}
\For{$u=1,2,\ldots,U_{train}$}
\State $\hat{\mathbf{r}}_{u} \gets
\Call{Multinomial}{\operatorname{softmax}(f_{out}(\mathbf{x}_{u}, \mathbf{z}_{u}, \mathbf{a}_{u}))}$
\EndFor
\State $L_{train} \gets \sum_{u=1}^{U_{train}} \ln p(\hat{\mathbf{r}}_{u})$,
\State Update weights $\boldsymbol{\psi}$ to maximize $L_{train}$
\EndFor
\State {\bfseries Until} $L_{val} \gets \sum_{u=U_{train}}^{U_{train}+U_{val}} \ln p(\hat{\mathbf{r}}_{u})$ decreases.
\For{$u=U-U_{test}, \ldots,U$} $\hat{\mathbf{r}}_{u} \gets f_{out}(\mathbf{x}_{u}, \mathbf{z}_{u},\mathbf{a}_{u})$
\EndFor
\State \Return $\{\hat{\mathbf{r}}_{u}\}_{u=U-U_{test}}^U$ as the estimated preferences.
\EndFunction
\State $\{\mathbf{z}_u\}_{u=1}^U$ $\gets$
\Call{ExposureNet}{$\{\mathbf{a}_{u}\}$}.
\State $\{\hat{\mathbf{r}}_{u}\}_{u=U-U_{test}}^{U}$ $\gets$
\Call{OutNet}{$\{\mathbf{r}_{u}\}$, $\{\mathbf{z}_{u}\}$, $\{\mathbf{x}_{u}\}$, $\{\mathbf{a}_{u}\}$}.
\For{$u=U-U_{test}, \ldots,U$} $\hat{\mathbf{r}}_{u}[\mathbf{a}_{u}==1] \gets - \inf$, $\operatorname{sort}(\hat{\mathbf{r}}_{u})$ and get top-$K$ items for recommendation.
\EndFor

\end{algorithmic}
\end{algorithm}

\subsection{Potential Rating Prediction for Recommendations}

\noindent Deep-Deconf is designed to predict unbiased ratings associated with any exposure vector $\mathbf{a}_{u} \in \{0, 1\}^{I}$, $\textit{i.e.,}$ $\mathbf{r}_{u}(\mathbf{a}_{u})$ with a low variance. However, which $\mathbf{a}_{u}$ should be selected for prediction is undetermined, as it demands the answer to the exact question we are trying to solve: which items should be exposed to users. Wang et al. \cite{wang2020causal} proposed to use $\mathbf{a}^{obs+K}_{u}$, \textit{i.e.,} the exposure of the originally observed items and $K$ newly recommended items, as the exposure to calculate the ratings for recommendations. However, for Deep-Deconf, enumerating every $\mathbf{r}_{u}(\mathbf{a}^{obs+K}_{u})$ and taking the expectation is clearly infeasible. Still, we propose an approximate strategy for the prediction. 
We assume that the number of recommended items $K$ (\textit{e.g.}, Top 20) is small compared with the size of the item pool (\textit{e.g.}, 10,000). Based on this assumption, the observed exposure $\mathbf{a}_{u}^{obs}$ can be used as a surrogate for $\mathbf{a}^{obs+K}_{u}$ to calculate $\mathbf{r}_{u}(\mathbf{a}^{obs+K}_{u})$ for predictions. 

Although with this prediction strategy, the same recommendations are made for users who have the same $\mathbf{a}_{u}^{obs}$, it does not mean that Deep-Deconf is non-personalized, because the pervasive unobserved confounders make the probability density of the exposures $\mathbf{a} _ u$ for each user concentrate on a small but unique niche of the exponentially large causal spaces if the number of exposed items exceed certain value (Actually, for a system with 5,000 items, the possible exposure combination of five items exceeds the population on earth). Therefore, $\mathbf{a} ^ {obs} _ u$ is \textit{per se} a very good representation of the user $u$ (although this can increase the estimand variance as discussed in Section \ref{seq:var_redu}). Moreover, a direct strategy to make Deep-Deconf "personalized" is to use the previous ratings of a user as extra user features in the outcome prediction model (they are pre-treatment variables because they remain unaffected by current exposures). However, experiments shows that this strategy only doubles the networks' trainable weights but improves hardly any performance. Therefore, we do not use the historical user ratings as pre-treatment variables.

\begin{figure*}
\centering
\centering
\includegraphics[width=0.98\textwidth]{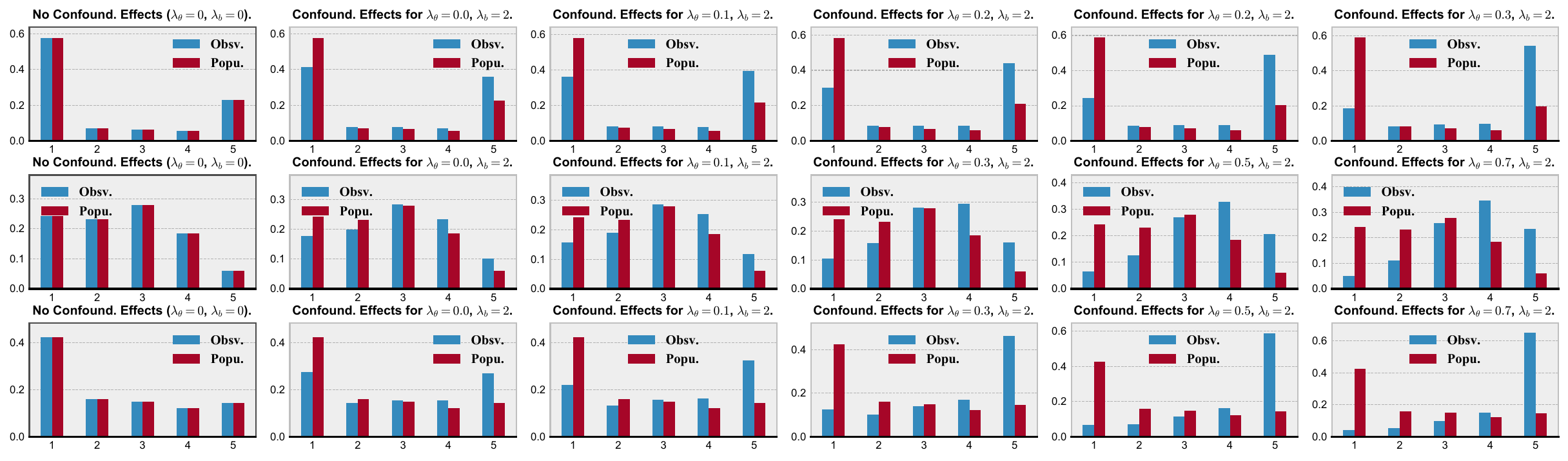}
\caption{Rating distribution before and after exposure under different confounding levels (Top: simulated dataset, Middle: \textit{ML-causal} dataset, Bottom: \textit{VG-causal} dataset; \textit{Obsv.}: observation, \textit{Popu.}: population). From this figure we can find that the discrepancy between the observed and true rating distributions becomes larger with the increase of confounding levels. (Detailed analysis of the established causal datasets can be referred to in Appendix \ref{seq:data_analysis})} 
\vspace{-3mm}
\label{fig:dataset_dist}
\end{figure*}

\section{Empirical Study}
\label{seq:emp}

\subsection{Datasets}
\label{seq:emp_data}

\subsubsection{Simulated Dataset}

\noindent We first conduct experiments on a simulated dataset. For each user, we draw a $K$-dimensional confounder $\mathbf{c}_{u}$ from $\mathcal{N}(\mathbf{0}, \mathbf{I}_{K})$ and calculate the user preference vector $\boldsymbol{\theta}_{u}$ conditional on the confounder $\mathbf{c}_{u}$ as $\boldsymbol{\theta}_{u} \sim \gamma_{\theta} \cdot \mathbf{c}_{u} + (1-\gamma_{\theta}) \cdot \mathcal{N}(\mathbf{0}, \mathbf{I}_{K})$. The item exposure rate $\alpha$ for the entire population is predetermined as 0.1. To simulate the exposure, we first generate a exposure propensity score for user $u$ as $\mathbf{a}_{u} = f_{a}(\mathbf{W}_{a} \mathbf{c}_{u})$. We then globally sort $\mathbf{a}_{u}$ for all users and set the top $\alpha$ items as the exposed items. The remaining part of the simulated dataset is generated as follows:
\begin{equation}
\begin{aligned} 
\mathbf{x}_{u} &\sim f_{PCA}(\boldsymbol{\theta}_{u} +  \boldsymbol{\epsilon}_{u}), \ \mathrm{where} \  {\boldsymbol{\epsilon}}_{u} \sim \mathcal{N}(0, \lambda_{u}^{-1}\mathbf{I}_{K});\\
\mathbf{r}_{u} &\sim  min(1+ Poisson(f_{r}(\mathbf{W}_{r} \boldsymbol{\theta}_{u} )), 5), \\
\end{aligned}
\end{equation}
\noindent where $\lambda_{b}$ and $\lambda_{\theta}$ control the strength of confounding effects, $\lambda_{u}$ controls the noise level of user features, and $\mathbf{W}_{r} \in \mathbb{R}^{K \times I}$ are the randomly initialized weights. Compared with the simulation procedure in \cite{wang2020causal}, we use functions $f_{a}$ and $f_{r}$ to simulate the non-linear influence of multi-cause confounders. In our experiment, we set both $f_{a}$ and $f_{r}$ as the ReLU function ($\mathrm{ReLU}(\mathbf{x})=\max(0, \mathbf{x})$), respectively.

\subsubsection{Real-world Datasets} 
\noindent Evaluating causal recommenders on real-world datasets faces great challenges, since usually we do not observe ratings for all the items of the users, and confounders make the model evaluations on a randomly split test set biased \cite{guo2021multisource}. Therefore, existing deep causal recommenders \cite{zou2020counterfactual} establish the causal datasets from real-world datasets.  \footnote{Note that an expensive and not user-friendly solution is to ask users to rate randomly-exposed items \cite{sato2020unbiased}. One such dataset is the Coat dataset established by \cite{louizos2017causal}. However, this dataset is too small in scale to train DNN-based recommenders (290 users and 300 items). Previously, the Yahoo R3 dataset \cite{marlin2009collaborative} can also be used to evaluate debiased recommenders. However, due to the current US export law, it is not publicly available. Therefore, these two datasets are usually not included to evaluate deep causal recommender systems.}

With a similar strategy, we create two semi real-world datasets, \textit{ML-causal} and \textit{VG-causal},  based on the  real-world Movielens-1m (ML-1m) \footnote{\url{https://grouplens.org/datasets/movielens/1m/}} and the Amazon-Videogames (Amazon-VG)\footnote{\url{https://jmcauley.ucsd.edu/data/amazon/}} datasets. In simulation, we train two VAEs with factorized logistic likelihood and multinomial likelihood on the exposures (binarized ratings) and ratings to get the corresponding generative distributions from user latent variables $\mathbf{z}_{\{a, r\}}$. The decoders of the exposure and rating VAEs are denoted as $f_{exp}$, $f_{rat}$, respectively. We then simulate a $K$-dimensional confounder $\mathbf{c}_{u}$ for each user from $\mathcal{N}(\mathbf{0}, \mathbf{I}_{K})$, and the user preference vector $\boldsymbol{\theta}_{u}$ conditional on the confounders is specified as $\boldsymbol{\theta}_{u} \sim \gamma_{\theta} \cdot \mathbf{c}_{u} + (1-\gamma_{\theta}) \cdot \mathcal{N}(\mathbf{0}, \mathbf{I}_{K})$; the constant  $\gamma_{\theta} \in [0,1]$ controls the strength of confounding. The user features are the noisy observation of her dimensional-reduced user preference vector $\operatorname{f_{PCA}}(\boldsymbol{\theta}_{u} + {\boldsymbol{\epsilon}}_{u})$, where $\boldsymbol{\epsilon}_{u} \sim \mathcal{N}(\mathbf{0}, \lambda_{u}^{-1} \mathbf{I}_{K})$. The exposure vector  $\mathbf{a}_{u}$ for user $u$ is generated from
\begin{equation}
\begin{aligned}
\mathbf{a}_{u} = \operatorname{argmin}_{\boldsymbol{\alpha}_{u} \in \{0, 1\}^{I}}\left(\sum _ {i} |\boldsymbol{\alpha}_{ui} - f_{exp}(\mathbf{c}_{u})_{i}|\right), \\ \ s.t. \ p_{casl}(A_{ui}=1)=p_{ori}(A_{ui}=1).
\end{aligned}
\end{equation}
The constraint is to ensure that the global item exposure rate of the causal datasets ($casl$) is the same as that of the original datasets ($ori$). Moreover, we define the set $\mathcal{R} = \{\mathbf{r_{u}} \in \mathbb{R}^{I} \mid r_{ui} \in \operatorname{range}(1,5) \}$ as the set of possible user rating vectors for $I$ items. The rating of user $u$ is generated from 
\begin{equation}
\begin{aligned}
\mathbf{r}_{u} = \operatorname{argmin}_{\boldsymbol{\gamma}_u \in \mathcal{R}}\left(\sum _ {i} |\boldsymbol{\gamma}_{ui} - f_{rat}(\boldsymbol{\theta}_{u} + \gamma_{b} * \mathbf{c}_{u})_{i}|\right), \\
\ s.t. \ p_{casl}(R_{ui}=r)=p_{ori}(R_{ui}=r), \ \forall r \in \operatorname{range}(1,5), 
\end{aligned}
\end{equation}
where $\gamma_{b}$ controls the strength of basic confounding level (since zero confounding is non-existent). The constraint ensures the global rating distribution in the causal datasets to be the same as the original datasets. The observed ratings $\mathbf{r}^{obs}_{u}$ is calculated by masking the original ratings with exposures, $\mathbf{r}^{obs}_{u} = \mathbf{r}_{u} \cdot \mathbf{a}_{u}$. The schematic illustration of the establishment can be referred to in Appendix \ref{sec:establish}. The global item popularity distribution before and after the exposure under different levels of confounding effect are shown in Fig. (\ref{fig:dataset_dist}). The statistics of the established simulated, ML-causal and, VG-causal datasets are in Table (\ref{tab:dataset}).

\begin{table}[]
\centering
\caption{Attributes of the established \textit{simulated}, \textit{ML-Causal} and \textit{VG-causal} datasets. In the table, \% density refers to the density of the
rating matrix, avg/std \#exp. refer to the corresponding statistics of the number of items that are exposed to the users.}
\label{tab:dataset}
\begin{tabular}{lcccc}

\toprule
dataset      & \#users & \#items & \%density & \#avg $\pm$ std exp.\\
\midrule
Simulated    & 4,000  & 2,000    & 1.000\% & 20 $\pm$ 11\\
ML-causal    & 6,000  & 3,706  & 4.468\%   & 165 $\pm$ 192\\
VG-causal    & 7,253  & 4,338  & 0.406\% &  17 $\pm$ 16 \\
\bottomrule
\end{tabular}
\end{table}

\subsection{Implementation Details and Training Strategy}

\label{sec:imple}

\noindent When establishing the causal datasets, the dimension $K$ of the user preference variables and the confounders is set to 100, 100, 100, respectively. The dimension of the user features is set to 10. The structure of the exposure and outcome models of Deep-Deconf is set to $\{I + \{F+K\} \rightarrow K \rightarrow I\}$, where $I$ is the number of items, $F$ is the dimension of user features, and $K$ is the latent dimension. The models are trained with Adam optimizer \cite{kingma2014adam}, with learning rate of $1e^{-3}$ for 100 epochs. For the exposure model, 20\% of the observed exposures of the validation users are hold-out for predictive check \cite{rubin1984bayesianly}. The selection of the outcome model generally follows the same procedure, where R@20 and N@20 on hold-out ratings are monitored as the metric. For all three datasets, we search $K_{s} \in \{50, 100, 150, 200\}$ and find that $K_{s}$ equals $K$ indeed achieves the best performance among models with all searched structures. 

\subsection{Evaluation Strategy}

\begin{table*}[t]
\small
\caption{Model comparisons under different confounding levels. The best method is highlighted in \textbf{boldface}. For each method, the best evaluated performance w.r.t. the confounding level is highlighted with {\color{blue}blue} and {\color{red}red} for R@20 and N@20, respectively.}
\label{tab:comp}
\centering
\begin{tabular}{lcccccccccc}
\hline
\multicolumn{1}{|l|}{Simulated} & \multicolumn{2}{c|}{$\lambda_{\theta}=0.1$}                   & \multicolumn{2}{c|}{$\lambda_{\theta}=0.3$}                   & \multicolumn{2}{c|}{$\lambda_{\theta}=0.5$}                   & \multicolumn{2}{c|}{$\lambda_{\theta}=0.7$}                   & \multicolumn{2}{c|}{$\lambda_{\theta}=0.9$}                   \\ \hline
\multicolumn{1}{|l|}{Methods}     & \multicolumn{1}{c|}{R@20}  & \multicolumn{1}{c|}{N@20}  & \multicolumn{1}{c|}{R@20}  & \multicolumn{1}{c|}{N@20}  & \multicolumn{1}{c|}{R@20}  & \multicolumn{1}{c|}{N@20}  & \multicolumn{1}{c|}{R@20}  & \multicolumn{1}{c|}{N@20}  & \multicolumn{1}{c|}{R@20}  & \multicolumn{1}{c|}{N@20}  \\ \hline
\multicolumn{1}{|l|}{WMF}         & \multicolumn{1}{c|}{0.572} & \multicolumn{1}{c|}{0.568} & \multicolumn{1}{c|}{0.569} & \multicolumn{1}{c|}{0.570} & \multicolumn{1}{c|}{0.575} & \multicolumn{1}{c|}{0.576} & \multicolumn{1}{c|}{\color{blue}0.595} & \multicolumn{1}{c|}{\color{red}{0.592}} & \multicolumn{1}{c|}{0.591} & \multicolumn{1}{c|}{0.589} \\
\multicolumn{1}{|l|}{IPW-MF}      & \multicolumn{1}{c|}{0.579} & \multicolumn{1}{c|}{0.573} & \multicolumn{1}{c|}{0.581} & \multicolumn{1}{c|}{0.582} & \multicolumn{1}{c|}{0.580} & \multicolumn{1}{c|}{0.586} & \multicolumn{1}{c|}{\color{blue}0.601} & \multicolumn{1}{c|}{ 0.595} & \multicolumn{1}{c|}{0.582} & \multicolumn{1}{c|}{\color{red}0.606} \\
\multicolumn{1}{|l|}{Deconf-MF}   & \multicolumn{1}{c|}{0.578} & \multicolumn{1}{c|}{0.580} & \multicolumn{1}{c|}{0.577} & \multicolumn{1}{c|}{0.592} & \multicolumn{1}{c|}{0.595} & \multicolumn{1}{c|}{0.593} & \multicolumn{1}{c|}{\color{blue} 0.612} & \multicolumn{1}{c|}{0.615} & \multicolumn{1}{c|}{0.609} & \multicolumn{1}{c|}{\color{red} 0.623} \\ \hline
\multicolumn{1}{|l|}{Concat-VAE}  & \multicolumn{1}{c|}{0.626} & \multicolumn{1}{c|}{0.634} & \multicolumn{1}{c|}{0.628} & \multicolumn{1}{c|}{0.653} & \multicolumn{1}{c|}{0.669} & \multicolumn{1}{c|}{0.685} & \multicolumn{1}{c|}{\color{blue}0.677} & \multicolumn{1}{c|}{\color{red}0.690} & \multicolumn{1}{c|}{0.671} & \multicolumn{1}{c|}{0.685} \\
\multicolumn{1}{|l|}{VSR-VAE}     & \multicolumn{1}{c|}{0.639} & \multicolumn{1}{c|}{0.655} & \multicolumn{1}{c|}{0.652} & \multicolumn{1}{c|}{\textbf{0.674}}      & \multicolumn{1}{c|}{0.678}      & \multicolumn{1}{c|}{0.693}      & \multicolumn{1}{c|}{\color{blue}0.685}      & \multicolumn{1}{c|}{\color{red}0.697}      &
\multicolumn{1}{c|}{0.680}  &\multicolumn{1}{c|}{0.694}      \\
\multicolumn{1}{|l|}{Deep-Deconf} & \multicolumn{1}{c|}{\textbf{0.650}} & \multicolumn{1}{c|}{\textbf{0.667}} & \multicolumn{1}{c|}{\textbf{0.664}} & \multicolumn{1}{c|}{0.672} & \multicolumn{1}{c|}{\textbf{0.691}} & \multicolumn{1}{c|}{\textbf{0.706}} & \multicolumn{1}{c|}{\color{blue}\textbf{0.696}} & \multicolumn{1}{c|}{\color{red}\textbf{0.712}} & \multicolumn{1}{c|}{\textbf{0.688}} & \multicolumn{1}{c|}{\textbf{0.703}} \\ \hline
\multicolumn{1}{|l|}{p-value} & \multicolumn{1}{c|}{1.70E-4} & \multicolumn{1}{c|}{3.89E-3} & \multicolumn{1}{c|}{3.65E-2} & \multicolumn{1}{c|}{$>0.1$} & \multicolumn{1}{c|}{8.72E-3} & \multicolumn{1}{c|}{2.03E-2} & \multicolumn{1}{c|}{1.91E-2} & \multicolumn{1}{c|}{7.12E-3} & \multicolumn{1}{c|}{$>0.05$} & \multicolumn{1}{c|}{4.36E-2} \\ \hline
                                  &                            &                            &                            &                            &                            &                            &                            &                            &                            &                            \\
\hline
\multicolumn{1}{|l|}{VG-causal} & \multicolumn{2}{c|}{$\lambda_{\theta}=0.1$}                   & \multicolumn{2}{c|}{$\lambda_{\theta}=0.3$}                   & \multicolumn{2}{c|}{$\lambda_{\theta}=0.5$}                   & \multicolumn{2}{c|}{$\lambda_{\theta}=0.7$}                   & \multicolumn{2}{c|}{$\lambda_{\theta}=0.9$}                   \\ \hline
\multicolumn{1}{|l|}{Methods}     & \multicolumn{1}{c|}{R@20}  & \multicolumn{1}{c|}{N@20}  & \multicolumn{1}{c|}{R@20}  & \multicolumn{1}{c|}{N@20}  & \multicolumn{1}{c|}{R@20}  & \multicolumn{1}{c|}{N@20}  & \multicolumn{1}{c|}{R@20}  & \multicolumn{1}{c|}{N@20}  & \multicolumn{1}{c|}{R@20}  & \multicolumn{1}{c|}{N@20}  \\ \hline
\multicolumn{1}{|l|}{WMF}         & \multicolumn{1}{c|}{0.318} & \multicolumn{1}{c|}{0.306} & \multicolumn{1}{c|}{\color{blue}0.325} & \multicolumn{1}{c|}{0.312} & \multicolumn{1}{c|}{0.321} & \multicolumn{1}{c|}{0.310} & \multicolumn{1}{c|}{0.322} & \multicolumn{1}{c|}{\color{red}{0.314}} & \multicolumn{1}{c|}{0.313} & \multicolumn{1}{c|}{0.301} \\
\multicolumn{1}{|l|}{IPW-MF}      & \multicolumn{1}{c|}{0.325} & \multicolumn{1}{c|}{0.319} & \multicolumn{1}{c|}{0.330} & \multicolumn{1}{c|}{0.324} & \multicolumn{1}{c|}{\color{blue}0.328} & \multicolumn{1}{c|}{0.319} & \multicolumn{1}{c|}{0.327} & \multicolumn{1}{c|}{\color{red} 0.321} & \multicolumn{1}{c|}{0.320} & \multicolumn{1}{c|}{0.314} \\
\multicolumn{1}{|l|}{Deconf-MF}   & \multicolumn{1}{c|}{0.333} & \multicolumn{1}{c|}{0.331} & \multicolumn{1}{c|}{0.341} & \multicolumn{1}{c|}{0.336} & \multicolumn{1}{c|}{0.340} & \multicolumn{1}{c|}{0.341} & \multicolumn{1}{c|}{\color{blue} 0.344} & \multicolumn{1}{c|}{\color{red} 0.346} & \multicolumn{1}{c|}{0.336} & \multicolumn{1}{c|}{0.339} \\ \hline
\multicolumn{1}{|l|}{Concat-VAE}  & \multicolumn{1}{c|}{0.369} & \multicolumn{1}{c|}{0.354} & \multicolumn{1}{c|}{0.385} & \multicolumn{1}{c|}{0.376} & \multicolumn{1}{c|}{\color{blue}0.405} & \multicolumn{1}{c|}{\color{red}0.398} & \multicolumn{1}{c|}{0.385} & \multicolumn{1}{c|}{0.378} & \multicolumn{1}{c|}{0.388} & \multicolumn{1}{c|}{0.379} \\
\multicolumn{1}{|l|}{VSR-VAE}     & \multicolumn{1}{c|}{0.377}      & \multicolumn{1}{c|}{0.372}      & \multicolumn{1}{c|}{0.398}      & \multicolumn{1}{c|}{0.390}      & \multicolumn{1}{c|}{\color{blue}0.422}      & \multicolumn{1}{c|}{\color{red}0.414}      & \multicolumn{1}{c|}{0.401}      & \multicolumn{1}{c|}{0.393}      & \multicolumn{1}{c|}{0.400}      & \multicolumn{1}{c|}{0.392}      \\
\multicolumn{1}{|l|}{Deep-Deconf} & \multicolumn{1}{c|}{\textbf{0.386}} & \multicolumn{1}{c|}{\textbf{0.379}} & \multicolumn{1}{c|}{\textbf{0.407}} & \multicolumn{1}{c|}{\textbf{0.399}} & \multicolumn{1}{c|}{\color{blue}\textbf{0.431}} & \multicolumn{1}{c|}{\color{red}\textbf{0.420}} & \multicolumn{1}{c|}{\textbf{0.410}} & \multicolumn{1}{c|}{\textbf{0.404}} & \multicolumn{1}{c|}{\textbf{0.414}} & \multicolumn{1}{c|}{\textbf{0.401}} \\ \hline
\multicolumn{1}{|l|}{p-value} & \multicolumn{1}{c|}{9.96E-5} & \multicolumn{1}{c|}{2.47E-3} & \multicolumn{1}{c|}{9.80E-3} & \multicolumn{1}{c|}{2.08E-2} & \multicolumn{1}{c|}{1.91E-3} & \multicolumn{1}{c|}{8.78E-3} & \multicolumn{1}{c|}{1.47E-2} & \multicolumn{1}{c|}{2.72E-3} & \multicolumn{1}{c|}{7.38E-3} & \multicolumn{1}{c|}{5.71E-3} \\ \hline
                                  &                            &                            &                            &                            &                            &                            &                            &                            &                            &                            \\ \hline
\multicolumn{1}{|l|}{ML-causal}   & \multicolumn{2}{c|}{$\lambda_{\theta}=0.1$}                   & \multicolumn{2}{c|}{$\lambda_{\theta}=0.3$}                                          & \multicolumn{2}{c|}{$\lambda_{\theta}=0.5$}                   & \multicolumn{2}{c|}{$\lambda_{\theta}=0.7$}                    & \multicolumn{2}{c|}{$\lambda_{\theta}=0.9$}                   \\ \hline
\multicolumn{1}{|l|}{Methods}     & \multicolumn{1}{c|}{R@20}  & \multicolumn{1}{c|}{N@20}  & \multicolumn{1}{c|}{R@20}  & \multicolumn{1}{c|}{N@20}                         & \multicolumn{1}{c|}{R@20}  & \multicolumn{1}{c|}{N@20}  & \multicolumn{1}{c|}{R@20}  & \multicolumn{1}{c|}{N@20}   & \multicolumn{1}{c|}{R@20}  & \multicolumn{1}{c|}{N@20}  \\ \hline
\multicolumn{1}{|l|}{WMF}         & \multicolumn{1}{c|}{0.068} & \multicolumn{1}{c|}{0.062} & \multicolumn{1}{c|}{0.072} & \multicolumn{1}{c|}{\color{red}0.071} & \multicolumn{1}{c|}{\color{blue}0.074} & \multicolumn{1}{c|}{0.070} & \multicolumn{1}{c|}{0.073} & \multicolumn{1}{c|}{0.068} & \multicolumn{1}{c|}{0.065} & \multicolumn{1}{c|}{0.059} \\
\multicolumn{1}{|l|}{IPW-MF}      & \multicolumn{1}{c|}{0.067} & \multicolumn{1}{c|}{0.062} & \multicolumn{1}{c|}{\color{blue}0.075} & \multicolumn{1}{c|}{\color{red}0.073}                        & \multicolumn{1}{c|}{0.066} & \multicolumn{1}{c|}{0.072} & \multicolumn{1}{c|}{0.071} & \multicolumn{1}{c|}{0.074}  & \multicolumn{1}{c|}{0.066} & \multicolumn{1}{c|}{0.065} \\
\multicolumn{1}{|l|}{Deconf-MF}   & \multicolumn{1}{c|}{0.076} & \multicolumn{1}{c|}{0.069} & \multicolumn{1}{c|}{\color{blue}0.084} & \multicolumn{1}{c|}{0.077}                        & \multicolumn{1}{c|}{0.082} & \multicolumn{1}{c|}{0.073} & \multicolumn{1}{c|}{0.080} & \multicolumn{1}{c|}{\color{red}0.079}  & \multicolumn{1}{c|}{0.078} & \multicolumn{1}{c|}{0.073} \\ \hline
\multicolumn{1}{|l|}{Concat-VAE}  & \multicolumn{1}{c|}{0.094} & \multicolumn{1}{c|}{0.097} & \multicolumn{1}{c|}{0.103} & \multicolumn{1}{c|}{0.106}                        & \multicolumn{1}{c|}{0.105} & \multicolumn{1}{c|}{0.101} & \multicolumn{1}{c|}{\color{blue}0.113} & \multicolumn{1}{c|}{\color{red}0.108}  & \multicolumn{1}{c|}{0.101} & \multicolumn{1}{c|}{0.103} \\
\multicolumn{1}{|l|}{VSR-VAE}     & \multicolumn{1}{c|}{0.103}      & \multicolumn{1}{c|}{0.102}      & \multicolumn{1}{c|}{0.109}      & \multicolumn{1}{c|}{\textbf{0.112}}                             & \multicolumn{1}{c|}{0.110}      & \multicolumn{1}{c|}{0.107}      & \multicolumn{1}{c|}{\color{blue}0.117}      & \multicolumn{1}{c|}{\color{red}0.114}       & \multicolumn{1}{c|}{0.106}      & \multicolumn{1}{c|}{0.106}      \\
\multicolumn{1}{|l|}{Deep-Deconf} & \multicolumn{1}{c|}{\textbf{0.108}} & \multicolumn{1}{c|}{\textbf{0.105}} & \multicolumn{1}{c|}{\textbf{0.113}} & \multicolumn{1}{c|}{0.111}                        & \multicolumn{1}{c|}{\textbf{0.117}} & \multicolumn{1}{c|}{\textbf{0.109}} & \multicolumn{1}{c|}{\color{blue}\textbf{0.124}} & \multicolumn{1}{c|}{\color{red}\textbf{0.121}}  & \multicolumn{1}{c|}{\textbf{0.115}} & \multicolumn{1}{c|}{\textbf{0.109}} \\ \hline
\multicolumn{1}{|l|}{p-value} & \multicolumn{1}{c|}{2.40E-4} & \multicolumn{1}{c|}{2.23E-2} & \multicolumn{1}{c|}{1.42E-3} & \multicolumn{1}{c|}{$> 0.1$} & \multicolumn{1}{c|}{2.96E-3} & \multicolumn{1}{c|}{$> 0.1$} & \multicolumn{1}{c|}{4.46E-4} & \multicolumn{1}{c|}{1.38E-3} & \multicolumn{1}{c|}{1.39E-4} & \multicolumn{1}{c|}{$>0.05$} \\ \hline
\end{tabular}
\end{table*}

\noindent We evaluate the model performance under strong generalization \cite{liang2015content}, where the observed item exposures and ratings for validation and test users are used only for inference purpose. When training the exposure model, we put aside 20\% of the observed exposures for predictive checks \cite{rubin1984bayesianly}, where the best model is selected by log-likelihood of the hold-out exposures. The outcome model is selected by how well it ranks the hold-out observed interactions for the validation users. The ranking quality is evaluated by R@$K$ and N@$K$, where  R@$K$ is the Top-$K$ recall. If we denote the item at ranking position $r$ by $i(r)$ and the set of hold-out items for the user by $\mathcal{I}_{u}$, R@$K$ is calculated as:
\begin{equation}
\operatorname{R} @ K(i)=\frac{\sum_{r=1}^{M} \mathbb{I}\left[j(r) \in \mathcal{I}_{u}\right]}{\min \left(M,\left|\mathcal{I}_{u}\right|\right)},
\end{equation}
\noindent where $\mathbb{I}$ in the numerator is the indicator function, and the denominator is the minimum of $K$ and the number of hold-out items. N@$K$ is the normalized DCG defined as follows:
\begin{equation}   \operatorname{DCG} @ K(u)=\sum_{r=1}^{K} \frac{2^{\mathbb{I}\left[i(r) \in \mathcal{I}_{u}\right]}-1}{\log (r+1)}.
\end{equation}
The model is selected by N@20 on validation users where 20\% of the observed ratings are hold-out for model evaluation. We choose only $K=20$ for both metrics to report because we found that the performance trend is similar for different $K$ in our experiments (we have tested $K \in range(5, 50, 5)$). The R@20 and N@20 on test users with fully observed ratings for all items averaged over five different splits of the datasets are reported as the unbiased model performance.

\subsection{Comparisons with Baselines}
\label{seq:emp_comp}

\subsubsection{Baselines} 
\noindent Our primary baseline is the deconfounded recommender (Deconf-MF) \cite{wang2020causal}\footnote{\url{https://github.com/yixinwang/causal-recsys-public}}, which also models the recommendation as a multiple causal inference problem. In Deconf-MF, the substitute confounders and ratings are estimated by linear Poisson factorization. Two other causality-based recommenders are both based on propensity-score re-weighting, which eliminates the confounding bias by re-weighting the ratings by the inverse of propensity scores \textit{i.e.,} the chance of their exposures conditional on covariates (previous exposures and user features). The first method is the inverse propensity weighting matrix factorization (IPW-MF) \cite{schnabel2016recommendations}, where the propensity scores are estimated by simple regression. Another method is the variational sample re-weigting (VSR) \cite{zou2020counterfactual}, which considers the multiple causes as a bundle treatment and estimates the propensity scores via latent variables instead of observations. For a fair comparison, the two matrix factorization-based methods are augmented with user features in an SVD++ manner \cite{koren2008factorization} and the VAE-based methods concatenates user features with exposures as the extra inputs, where improvement has been observed compared to their original forms. The non-causal baselines we include are the weighted matrix factorization \cite{hu2008collaborative} (WMF, also augmented with user features) and the concat-VAE, which is a variant of Deep-Deconf where the exposure model is removed to demonstrate the effectiveness of causal reasoning.

\subsubsection{Experimental Setups} 

\noindent In simulation, we fix the basic confounding level $\gamma_{b}$ to 2.0 as with the empirical estimation of Wang et al. \cite{wang2020causal}, and we vary the strength of user preference confounding effects by changing $\gamma_{\theta} \in \operatorname{range}(0.1,0.2,0.9)$. The models are then evaluated under different confounding levels.

\subsubsection{Comparisons with the State-of-the-Art} \noindent The comparison results are demonstrated in Table (\ref{tab:comp}). From Table (\ref{tab:comp}) we can find that the vanilla WMF performs the worst among all the methods that we draw comparisons with, especially when the ratings are heavily confounded. IPW-MF addresses the confounding bias by re-weighting the ratings by the propensity scores estimated through regression. While improvement has been observed over WMF in most cases, the unbiasedness of IPW-MF requires two strong assumptions, \textit{i.e.,} unconfoundedness and a correctly specified propensity model, which relies heavily on the expertise of the researchers. Therefore, it hinders IPW-MF's further improvement. Deconf-MF weakens the unconfoundedness assumption of IPW-MF to the non-existence of single-cause confounders, and is the best matrix factorization-based baselines demonstrated in the middle part of three sub-tables of Table (\ref{tab:comp}). However, since Deconf-MF models both the exposures and the ratings as linear Poisson matrix factorization, it fails to capture non-linear influences of unobserved confounders. Moreover, it treats a user's ratings to different items separately in the outcome model, where co-recommendation effects cannot be considered to further improve recommendation performance. 

Consequently, Deconf-MF is outperformed by the Multi-VAE-based deep generative models, even if Concat-VAE is not causality-based and VSR is based on propensity score-reweighting which requires a stronger unconfoundedness assumption for unbiasedness to hold. Combining the advantages of non-linear collaborative modeling ability of Multi-VAE and model-agnostic debiasing advantage of the substitute confounder-based causal inference for recommendations, Deep-Deconf achieves a systematic performance improvement compared with Deconf-MF while being more robust than CondVAE and VSR faced with unobserved confounders. Therefore, the experiments demonstrate the superiority of the Deep-Deconf to other causal recommenders.

\subsection{Duality of Multi-Cause Confounders} 

\noindent From Table (\ref{tab:comp}) we can find that the best results for each method w.r.t. the confounding level, which are marked with colors blue and red for $R@20$ and $N@20$, respectively, appear in the middle of the Table. This shows an interesting phenomenon that the performance for all the methods improves first and then deteriorates with the increase of confounding level. This is against the naive intuition that the recommendation performance should reduce monotonically when the confounding level increases. The phenomenon can also be discovered from Fig. 1 of \cite{wang2020causal}, but the authors provided no explanation to why it occurs. 

\subsubsection{Formulation of the Theory}
\noindent We propose the "duality of multi-cause confounders" to explain such a phenomenon. When the confounding level is low, users tend to consume items at random where item co-occurrences contains little collaborative information. Therefore, the models waste parameters to fit uninformative random item exposures, which degenerates the model performances. However, the observed rating distribution is a more faithful representation of the population rating distribution. This reduces the confounding bias. When the strength of confounding effects increases, although the co-occurrence of items demonstrates more regular patterns, the observed rating distribution deviates further from the true population distribution. Thus, the influence of confounding bias outweighs the introduction of extra item collaborative information and reduces the performance. 

In essence, the duality of multi-cause confounders results from their commonality among items, \textit{i.e.,} they are also shared item attributes.  Therefore, in contrast to their role in traditional single-cause inference problems, the confounders in recommendations, like \textit{yin} and \textit{yang}, exert their force in two opposite ways: On the one hand, since these confounders tend to be "shared" items attributes, they help explain why certain items tend to appear together and why certain items have never occurred at the same time. This introduces item collaborative information that is conducive to the recommendation of new items that are similar to the items that the user has interacted with. On the other hand, when the confounding effects are overly strong, the observed rating distribution diverges drastically from the true rating distribution, which leads to systematic bias due to the unbalanced representation of items in the data that severely degenerates the recommendation performance. (Appendix \ref{sec:app_duality}). 

\subsubsection{Discussions on Deconfounding}
\noindent The duality of multi-cause confounders naturally leads to a further research question: In the battle of the benefits and drawbacks of multi-cause confounders, which one will prevail? The answer, which the authors believe, is that no matter the results, a deconfounded recommender should always be preferred over a non-causality-based one if unobserved confounders indeed pervasively exist to affect the ratings, because greedily exploiting the collaborative information associated with unobserved confounders come at a price: It also inherits the confounding bias in the historical ratings, which hinders the further improvement of the recommendation model. Consider the extreme case where the exposures are entirely due to previous recommendations. In this case, the new recommender system will perform no better than the previous one by greedily exploiting. Deconfounded recommenders remedy the bias by balancing the under- and over- represented items with either sample re-weighting or controlling substitute confounders. Through these mechanisms, the recommender system can provide a more bias-free estimate of user preferences. Furthermore, the above analysis naturally leads to another conclusion that a good strategy to reduce the confounding bias from the data collection perspective is to avoid greedy exploitation and to add random exploration to the recommendation policy that collects the user ratings, where the randomness negates the influence of unobserved confounders. This is interesting, as it provides another justification for exploration, i.e., a commonly adopted strategy in reinforcement learning-based recommender systems, from a causal perspective. A brief discussion of the relationship between \textit{exploration} and \textit{deconfounding} can be referred to in Appendix \ref{sec:deconf}.

\subsection{Sensitivity Analysis to User Features}

\noindent We have proved that introducing user features as pre-treatment variables can reduce the estimand variance for Deep-Deconf as long as they are informative to rating prediction. However, how does the "informative level" of user features influence the estimand variance is unclear.  Recall that in our simulation, user features are generated by setting $\mathbf{x}_{u} = \operatorname{f_{PCA}}(\boldsymbol{\theta}_{u} + {\boldsymbol{\epsilon}}_{u})$, where $\boldsymbol{\theta}_{u}$ is the user preference and $\boldsymbol{\epsilon}_{u} \sim \mathcal{N}(\mathbf{0}, \lambda_{u}^{-1} \mathbf{I}_{K})$ is a random Gaussian noise. We control the informative level of the user features by setting $\lambda_{u} ^ {-1} \in \{0.1, 0.5, 0.9\}$ and evaluate Deep-Deconf. The performances are compared to a baseline where no user features are used. The results are summarized in Table (\ref{tab:noise}). From Table (\ref{tab:noise}) we can find that user features that are more informative to rating prediction (i.e., with less noise) indeed lead to a lower estimand variance, which is reflected by a higher performance when the ratings associated with one specific item exposure is extremely sparse. Moreover, the performance improvement is more significant when the simulated confounding level is high. When the user features are highly noisy ($\lambda_{f} = 0.9$), however, the model overfits on the noise and the performance degenerates to be similar to the baseline model where no user features are used.

\begin{table}[]
\small
\centering
\caption{Performance of Deep-Deconf when user features have different noise levels (N.F. means no user features)}
\label{tab:noise}
\begin{tabular}{lcccc}
\hline
\multicolumn{1}{|l|}{Simulated} & \multicolumn{2}{c|}{Low Conf. ($\lambda_{b}=0.3$)}                                & \multicolumn{2}{c|}{High Conf. ($\lambda_{b}=0.7$)}                               \\ \hline
\multicolumn{1}{|l|}{Noise}     & \multicolumn{1}{c|}{R@20}            & \multicolumn{1}{c|}{N@20}            & \multicolumn{1}{c|}{R@20}            & \multicolumn{1}{c|}{N@20}            \\ \hline
\multicolumn{1}{|l|}{0.1}       & \multicolumn{1}{c|}{\textbf{0.6665}} & \multicolumn{1}{c|}{\textbf{0.6726}} & \multicolumn{1}{c|}{\textbf{0.6967}} & \multicolumn{1}{c|}{\textbf{0.7127}} \\
\multicolumn{1}{|l|}{0.5}       & \multicolumn{1}{c|}{0.6602}          & \multicolumn{1}{c|}{0.6650}          & \multicolumn{1}{c|}{0.6934}          & \multicolumn{1}{c|}{0.7023}         \\
\multicolumn{1}{|l|}{0.9}       & \multicolumn{1}{c|}{0.6574}          & \multicolumn{1}{c|}{0.6518}          & \multicolumn{1}{c|}{0.6901}          & \multicolumn{1}{c|}{0.6887}          \\
\multicolumn{1}{|l|}{N.F.}      & \multicolumn{1}{c|}{0.6437}          & \multicolumn{1}{c|}{0.6594}          & \multicolumn{1}{c|}{0.6832}          & \multicolumn{1}{c|}{0.6686}          \\ \hline
                                &                                      &                                      &                                      &                                      \\ \hline
\multicolumn{1}{|l|}{VG-causal} & \multicolumn{2}{c|}{Low Conf. ($\lambda_{b}=0.3$)}                                & \multicolumn{2}{c|}{High Conf. ($\lambda_{b}=0.7$)}                               \\ \hline
\multicolumn{1}{|l|}{Noise}     & \multicolumn{1}{c|}{R@20}            & \multicolumn{1}{c|}{N@20}            & \multicolumn{1}{c|}{R@20}            & \multicolumn{1}{c|}{N@20}            \\ \hline
\multicolumn{1}{|l|}{0.1}       & \multicolumn{1}{c|}{\textbf{0.4071}} & \multicolumn{1}{c|}{\textbf{0.3974}} & \multicolumn{1}{c|}{\textbf{0.4098}} & \multicolumn{1}{c|}{\textbf{0.4039}} \\
\multicolumn{1}{|l|}{0.5}       & \multicolumn{1}{c|}{0.4034}          & \multicolumn{1}{c|}{0.3963}          & \multicolumn{1}{c|}{0.4036}          & \multicolumn{1}{c|}{0.3982}          \\
\multicolumn{1}{|l|}{0.9}       & \multicolumn{1}{c|}{0.4010}          & \multicolumn{1}{c|}{0.3928}          & \multicolumn{1}{c|}{0.3993}          & \multicolumn{1}{c|}{0.3934}          \\
\multicolumn{1}{|l|}{N.F.}      & \multicolumn{1}{c|}{0.3982}          & \multicolumn{1}{c|}{0.3909}          & \multicolumn{1}{c|}{0.3948}          & \multicolumn{1}{c|}{0.3912}          \\ \hline
                                &                                      &                                      &                                      &                                      \\ \hline
\multicolumn{1}{|l|}{ML-causal} & \multicolumn{2}{c|}{Low Conf. ($\lambda_{b}=0.3$)}                                & \multicolumn{2}{c|}{High Conf. ($\lambda_{b}=0.7$)}                               \\ \hline
\multicolumn{1}{|l|}{Noise}     & \multicolumn{1}{c|}{R@20}            & \multicolumn{1}{c|}{N@20}            & \multicolumn{1}{c|}{R@20}            & \multicolumn{1}{c|}{N@20}            \\ \hline
\multicolumn{1}{|l|}{0.1}       & \multicolumn{1}{c|}{\textbf{0.1130}} & \multicolumn{1}{c|}{\textbf{0.1108}} & \multicolumn{1}{c|}{\textbf{0.1240}} & \multicolumn{1}{c|}{\textbf{0.1212}}          \\
\multicolumn{1}{|l|}{0.5}       & \multicolumn{1}{c|}{0.1119}          & \multicolumn{1}{c|}{0.1094}          & \multicolumn{1}{c|}{0.1238}          & \multicolumn{1}{c|}{0.1201}          \\
\multicolumn{1}{|l|}{0.9}       & \multicolumn{1}{c|}{0.1088}          & \multicolumn{1}{c|}{0.1082}          & \multicolumn{1}{c|}{0.1203}          & \multicolumn{1}{c|}{0.1175}          \\
\multicolumn{1}{|l|}{N.F.}      & \multicolumn{1}{c|}{0.1092}          & \multicolumn{1}{c|}{0.1091}          & \multicolumn{1}{c|}{0.1172}          & \multicolumn{1}{c|}{0.1144}          \\ \hline
\end{tabular}
\end{table}

\section{Conclusions}
\label{seq:conc}

\noindent In this article, we proposed an effective deep factor model-based causal inference algorithm, Deep-Deconf, for recommender systems. By controlling substitute confounders inferred through factorized logistic VAE that render the observed exposures randomized Bernoulli trials, Deep-Deconf lower the multi-cause confounding bias, leading to more faithful estimation of user preferences. Moreover, we have proved that the variance of the estimated unbiased ratings can be substantially decreased by introducing user features as pre-treatment variables. We note that our algorithm can be plugged into any user-oriented auto-encoder-based recommender systems by adding a decoder branch that constrains the user-latent variable to generate factorized exposures. Therefore, we speculate that these models can also benefit from the confounding reduction advantage of our method with a modest extra computational overhead.

\appendix
\renewcommand*\appendixpagename{\Large Appendices}
\appendixpage

\section{Theoretical analysis of Deep-Deconf}

\subsection{Proof of Conditional Independence}

\noindent In the exposure model of Deep-Deconf, we claim that a decoder that takes substitute confounder $\mathbf{z}_{u}$ as input and reconstructs the exposure $\mathbf{a}_{u}$ with factorized logistic likelihood can renders $\mathbf{a}_{u}$ conditionally independent. Some may question that this is not possible since the weights that predict $a_{ui}$ for different $i$ are shared. This is not true. The fact is that the exposures cannot be marginally independent, as they are all governed by unobserved confounders. However, conditional on $\mathbf{z} _ u$, they can be independent, and generation of $\mathbf{a} _ u$ from $\mathbf{z} _ u$ can also be implemented via a shared decoder. The reason is as follows. For user $u$ with latent confounder $\mathbf{z} _ u$, if we denote the input vector to the last layer of the exposure network as $f(\mathbf{z} _ u)$, the logit of the exposure to item $i$ is calculated as $logit \ p(a_{ui}\mid \mathbf{z} _ u) = [\mathbf{W} f(\mathbf{z} _ u)] _ i =  \mathbf{w} _ i  \cdot f(\mathbf{z} _ u)$. Since $f(\mathbf{z} _ u)$ only depend on $\mathbf{z} _ u$, if the row vectors of the last layer weights $\mathbf{W}$ are independent, $\mathbf{z} _ u$ contains all information for $a_{ui}$ contained in $a_{uj}$. Therefore, $a_{ui}$ is conditionally independent of $a_{uj}$ given $\mathbf{z} _ u$, even if the weights used to infer different $a_{ui}$ are shared among all items. 

\subsection{Proof of Variance Reduction via User Features as Pre-treatment Variables}
\label{sec:proof_var_appendix}

\begin{figure}[t]
\centering
\includegraphics[width=0.47\textwidth]{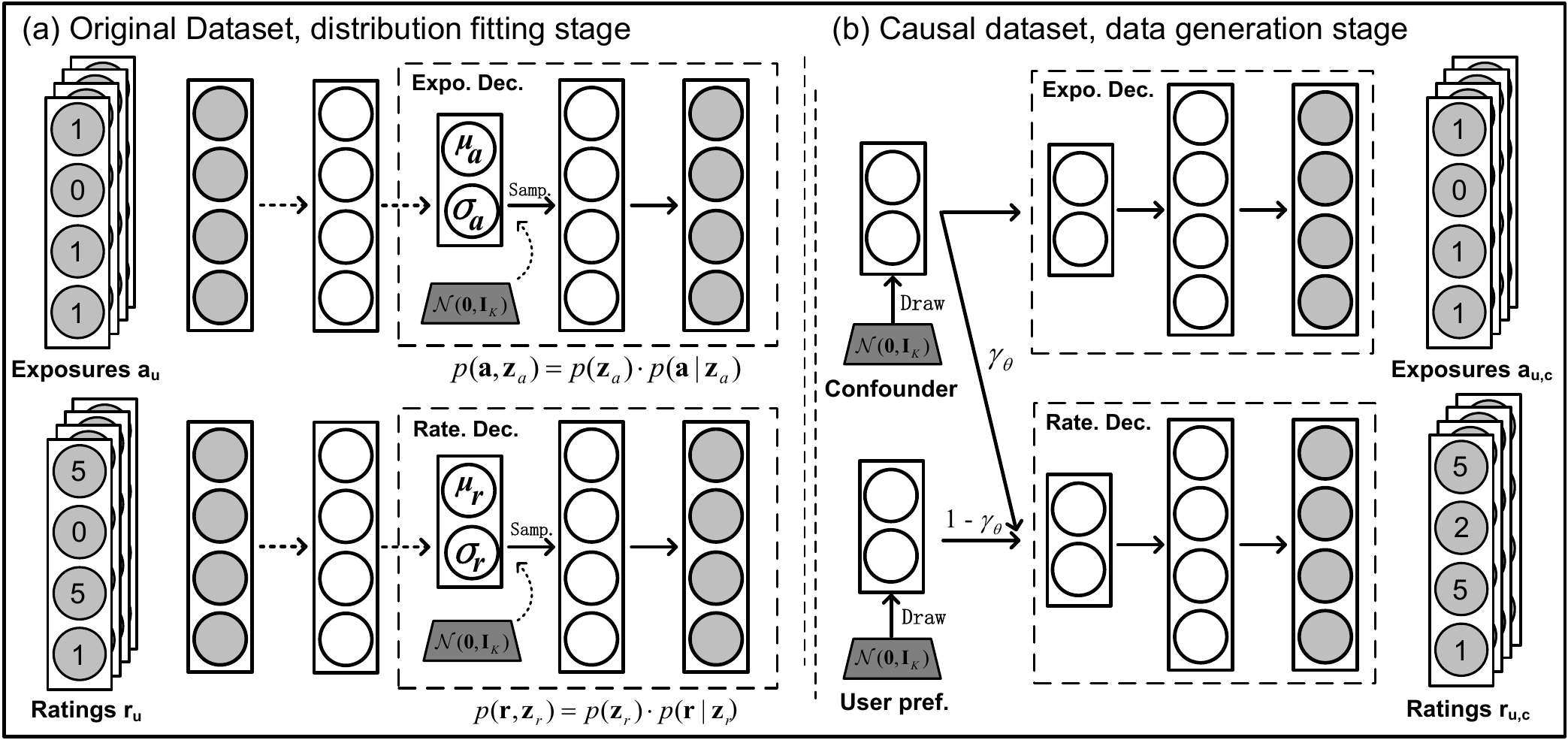}
\caption{A simplified illustration of the distribution fitting process and the confounding simulation process to create the real-world causal datasets used in experiments.}
\label{fig:data_est}
\end{figure}

\noindent In this section, we derive the sampling variance of the coefficient of exposure indicator obtained by ordinary least square (OLS) estimator before and after the introduction of user features as pre-treatment variables. Recall that after we simplify the network weight $\mathbf{W}$ to a diagonal matrix $\mathbf{w} \cdot \mathbf{I}$, user features and substitute confounders to scalars, the single-layer outcome prediction network for the ratings before the introduction of user features as pre-treatment variables becomes
\begin{equation}
\label{eq:before_pre}
        r_{ui}(a_{ui}) = w^{z}z_{u}+w_{i}^{a}a_{ui} + \alpha_{i} + \epsilon_{ui}.
\end{equation}
The OLS estimators for the coefficients can be specified as
\begin{equation}
    \left(\hat{w}^{z}, \hat{w}_{i}^{a}, \hat{\alpha}_{i}\right)=\operatorname{argmin} _{w^{z}, w_{i}^{a}, \alpha_{i}} \sum_{u=1}^{U}\left(r_{ui}-w^{z}z_{u}+w_{i}^{a}a_{ui} + \alpha_{i}\right)^{2}.
\end{equation}
As we have demonstrated in the main paper, since conditional on $z_{u}$, the ratings $r_{ui}$ can be viewed as generated from randomized experiments, $\hat{w}_{i}^{a}$ is an unbiased estimator for the population conditional average exposure effects of item $i$ on rating $r_{ui}$. The SUTVA assumption ensures the non-interference of exposures of different users, and the homoskedasticity assumption ensures that the variance does not vary by the change of $a_{ui}$ and $z_{u}$ \cite{imbens2015causal}. If we further assume $\mathbb{E}[\epsilon_{ui} \mid A_{ui}, Z_{u}]=0$ (the above assumptions are known as the Gaussian-Markov assumption), according to the Gauss theorem \cite{Johnson2003}, the sampling variance of $\hat{w}_{i}^{a}$ is
\begin{equation}
\label{eq:var_before}
    \hat{V}_{i}=\frac{\hat{\sigma}_{R_{ui} \mid A_{ui},Z_{u}}^{2}}{\sum_{u=1}^{U}\left(a_{ui}-\bar{a}_{ui}\right)^{2}}=s^{2} \cdot\left(\frac{1}{U_{i}^{1}}+\frac{1}{U_{i}^{0}}\right),
\end{equation}
\noindent where $s^{2} = \hat{\sigma}_{R_{ui} \mid A_{ui},Z_{u}}^{2}$ is the OLS variance of $R_{ui}$, $\bar{a}_{ui} = \sum_{u}a_{ui} / U$ is the average exposure number of item $i$ in the finite sample, $U_{i}^{1}$ is the  exposure count of item $i$ and $U_{i}^{0} = U-U_{i}^{1}$. The second equality can be derived with simple algebra based on the fact that $a_{ui} \in \{0, 1\}$ and therefore $a_{ui}^{2} = a_{ui}$. The $\hat{\sigma}_{R_{ui} \mid A_{ui},Z_{u}}^{2}$ can be calculated as the common variance across the two potential
outcome distributions,
\begin{equation}
\begin{aligned}
    \hat{\sigma}_{R_{ui} \mid A_{ui}, Z_{ui}}^{2}=\frac{1}{U-2}\Big(&\sum_{u: a_{ui}=1}\left(r_{ui}(1)-\bar{r}_{ui}(1)\right)^{2}\\+&\sum_{u: a_{ui}=0}\left(r_{ui}(0)-\bar{r}_{ui}(0)\right)^{2}\Big).
\end{aligned}
\end{equation}
\noindent After we multiply and divide $U$ on the R.H.S. of Eq. (\ref{eq:var_before}), we have $\hat{V}_{i} \overset{p}{\to} \frac{\sigma_{R_{ui} \mid Z_{ui},A_{ui}}^{2}}{U \cdot p_{i} \cdot(1-p_{i})}$ as $U$ approaches infinity, where $p_{i} = \operatorname{lim}^{p}_{U \to \inf} U^{1}_{i}/U$ is the population probability of the exposure of item $i$; this finishes our deduction of estimand variance before introducing the user features as pre-treatment variables. Post the introduction of the user features $x_{u}$, the outcome prediction network becomes
\begin{equation}
\label{eq:after}
        r_{ui}(a_{ui}) = w^{z}z_{u}+w_{i}^{a}a_{ui} + w^{x}x_{u} + \alpha_{i} + \epsilon_{ui}.
\end{equation}
\noindent Supposing again the variance does not vary by the change of the treatment indicator $a_{ui}$, the surrogate confounder $z_{u}$, and the user feature $x_{u}$ (\textit{i.e.,} homoscedasticity), with the SUTVA assumption and the assumption of $\mathbb{E}[\epsilon_{ui} \mid A_{ui}, Z_{u}, X_{u}]=0$, the limiting sampling variance of the OLS estimator for Eq. (\ref{eq:after}), \textit{i.e.,} $\hat{V}^{new}_{i}$, given the general case of $\hat{V}_{i}$, can be directly calculated as $\frac{\sigma_{R_{ui} \mid X_{u}, Z_{ui},A_{ui}}^{2}}{U \cdot p_{i} \cdot(1-p_{i})}$. By comparing the form of $\hat{V}^{new}_{i}$ and $\hat{V}_{i}$ we can find that the only difference lies in the OLS variance term in the numerator: if the features of a user $x_{u}$ are indicative to the prediction of her ratings $r_{ui}$, $\sigma_{R_{ui} \mid X_{u}, Z_{ui},A_{ui}}^{2}$ reduces considerably compared with $\sigma_{R_{ui} \mid Z_{ui},A_{ui}}^{2}$, which leads to a substantial decrease of estimand variance. This is especially favorable in our multiple causal case where the low sample efficiency due to large causal space hinders a precise inference.

\begin{figure}
\centering
\includegraphics[width=0.45\textwidth]{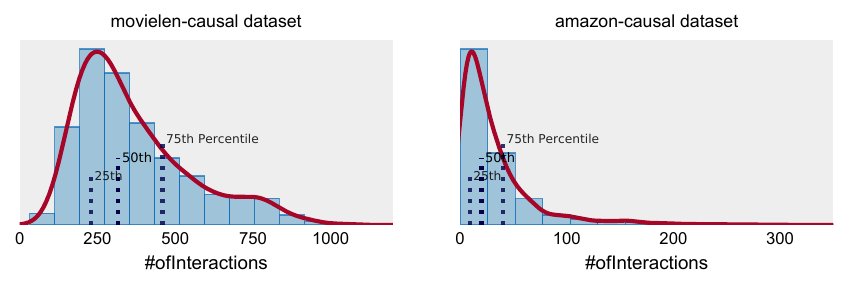}
\caption{Item popularity distribution of the established ML-causal and VG-causal datasets.}
\label{fig:pop_dist}
\end{figure}

\subsection{Discussion of Model Identifiability}

\noindent The authors have noticed that the deconfounder algorithm \cite{wang2019blessings}, based on which we designed the Deep-Deconf, has raised some discussions recently among researchers, including \cite{d2019multi}, \cite{ogburn2019comment}. Wang \& Blei have also responded to the comments in \cite{wang2019blessings}, \cite{wang2020towards}. The questions of \cite{d2019multi} were solved in \cite{wang2020causal} (see footnotes on Page 2), which stated that the focus of Deconf-MF is on estimating the expected potential outcome (ratings) if $K$ extra items are exposed (\textit{i.e.,} top-K recommendations), so it has different assumptions with counterexamples that rely on do-operators. The disagreement between Wang \& Blei and Ogburn et al. mainly lies in the assumptions required for the model identifiability. Ogburn et al. believed that Deconfounder requires extra assumptions such as the inferred substitute confounder $\mathbf{z}$ does not pick up post-treatment variables, etc. However, Wang et al. responded that these assumptions do not satisfy the requirement $\mathbf{z}$ can be pinpointed (see Theorem 7 in \cite{wang2020causal}). Similarly, the identifiability of Deep-Deconf also relies on the pinpoint requirement of the substitute confounders.

\section{Experiments on Real-world Datasets}

\subsection{Schematic Illustration of Dataset Establishment}
\label{sec:establish}

\noindent A schematic illustration for the establishment of the  real-world causal datasets is shown in Fig. (\ref{fig:data_est}).

\begin{figure}
\centering
\includegraphics[width=0.48\textwidth]{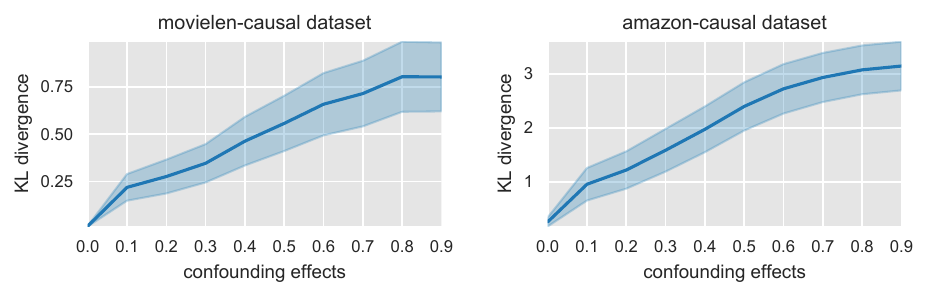}
\caption{Averaged KL-divergence between individual rating distributions before and after exposure.}
\label{fig:ind_rat_dist}
\end{figure}

\subsection{Item Popularity Distributions}

\noindent The popularity of an item is defined as the number of users who have rated that item. Since the item popularity distribution on the ML-causal and VG-causal datasets only depends on the exposure, we visualize the distribution with an arbitrary confounding effect. The results are illustrated in Fig.  (\ref{fig:pop_dist}). From Fig. (\ref{fig:pop_dist}) we can discover that the item popularity distribution of both the ML-causal and VG-causal datasets exhibits both long-tail and right-skewed characteristics, which faithfully reflects the item popularity distributions in the real-world scenario. 

\subsection{Rating Distribution Pre- and Post- Exposure}
\label{seq:data_analysis}
\noindent The global rating distribution of a recommendation dataset is defined as $P(R=r) \propto \sum_{u,i} \mathbf{1}(r_{ui}=r), \ \forall r \in \{1,2,3,4,5\}$, where $r=0$ denotes the rating is unobserved is excluded from consideration. If no confounder exists, the exposure matrix $\mathbf{A}_{rand}$ is a random matrix where the elements in $\mathbf{A}_{rand}$ are independent Bernoulli variables. The individual rating distribution for a user $P_{u}(R=r)$ can be defined accordingly. Since the observed ratings $\mathbf{R}^{obs}_{rand}$ is acquired by $\mathbf{R}^{obs}_{rand} = \mathbf{R}_{all} \times \mathbf{A}_{rand}$, the element-wise independence of $\mathbf{A}_{rand}$ ensures that the global and individual rating distributions of the observed ratings $\mathbf{R}^{obs}_{rand}$ is unbiased estimators to those of the population ratings $\mathbf{R}_{all}$. Unobserved confounders, however, create a spurious dependence of $\mathbf{R}^{obs}_{conf}$ on $\mathbf{A}_{conf}$, and therefore lead to systematic bias of the rating distributions in $\mathbf{R}^{obs}_{conf}$ after the exposure. Although in practice, the population rating matrix $\mathbf{R}_{all}$ is unobtainable, in our simulation, we have the users' ratings for all items (although only exposed ratings are visible to the algorithm for training). Therefore, in this paper, we can visualize the confounders' effect on both global and individual rating distributions after exposure under various levels of confounding effects. 

We set $\lambda_{b}=0$ and $\lambda_{\theta}=0$ where no confounding effect exists and then fix $\lambda_{b}$  to 2, and vary $\lambda_{\theta}$ from 0.1 to 0.9. The comparisons of global rating distribution are shown in Fig. (\ref{fig:dataset_dist}). From Fig. (\ref{fig:dataset_dist}), we can find that an obvious observation of the confounding effect is that positively rated items are more likely to be exposed than their negatively rated counterparts, and the stronger the confounding effects are, the more unbalanced the exposure of highly-rated and lowly-rated items. This is quite interesting, since all we have done to simulate the confounders is to (a) train two VAEs to model the exposure and rating distributions of the ML-1m and Amazon-VG datasets and (b) make the latent variables that generate the exposure and ratings for a user correlated by taking a weighted sum of the confounder and user variable. This phenomenon also has a real-world explanation: Users tend to rate items they like and ignore items they dislike, which leads to systematic bias due to the gross  under-representation of items with negative ratings. Furthermore, we visualize the difference of the individual rating distributions before and after exposure under various levels of confounding effects. The averaged KL-divergence between true and observed individual rating distribution is in Fig. (\ref{fig:ind_rat_dist}).

\begin{figure}
\centering
\includegraphics[width=0.48\textwidth]{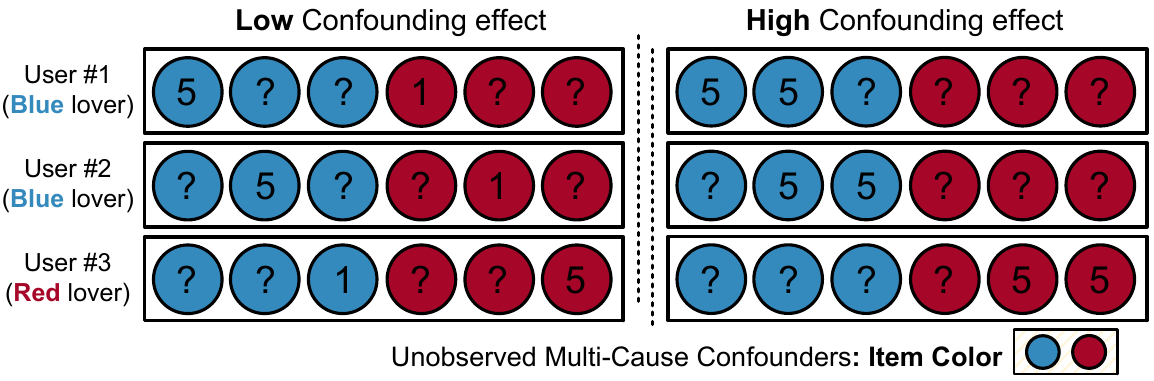}
\caption{A simple example to explain the duality of multi-cause confounders in recommendation systems.}
\label{fig:yinyang}
\end{figure}

\subsection{More on Duality of Multi-cause Confounders}
\label{sec:app_duality}

\noindent In the experiments, we have discovered that the recommendation quality improves first and then degenerates with the increase of confounding level. We conclude that "multi-cause confounders in recommender systems like \textit{yin} and \textit{yang}, exert their forces in two opposite ways." In this section, we provide a simple but intuitive example to further support the claim. Suppose that in our toy system, a subset of users is "blue lovers" who rate all blue items five and all red items one —similarly, a subset of "red lovers" rate just the opposite way. Moreover, blue lovers tend to be recommended with blue items, and red lovers with red items. Since color affects both the exposure and the rating of an item, item color is a confounder in the system. In addition, it is a multi-cause confounder because color is an attribute that are shared among all the items.

The observed ratings for the two blue lovers and one red lover under low and high confounding levels are illustrated in Fig. (\ref{fig:yinyang}). The left part of Fig. (\ref{fig:yinyang}) shows observations under no confounding effects. In such a case, the exposure probability of an item is independent of its color (1/3 for both red items and blue items), and the observed global and individual rating distributions $p(r=5) = p(r=1) = 1/2$ exactly matches the population rating distributions. However, since the item co-occurrences are random, no item collaborative information can be utilized to recommend new items with similar rating patterns. In contrast, the right part of Fig. (\ref{fig:yinyang}) shows rating observations with a high confounding level, where the exposure probability of an item clearly depends on its color and the user's preference. Under this circumstance, new items with similar user rating patterns with the items have already interacted by the users can be readily recommended based on item collaborative information (\textit{e.g.,} item \#3 to user \#1, and item \#1 to user \#2 based on their similar rating pattern to item \#2 for both users). However, the observed rating distribution (all are five) deviates drastically from the population distribution, which introduces a systematic bias between the true and observed distribution that could degrade the model performance.

\subsection{Exploration and Deconfounding}
\label{sec:deconf}

\begin{figure}
\centering
\includegraphics[width=0.48\textwidth]{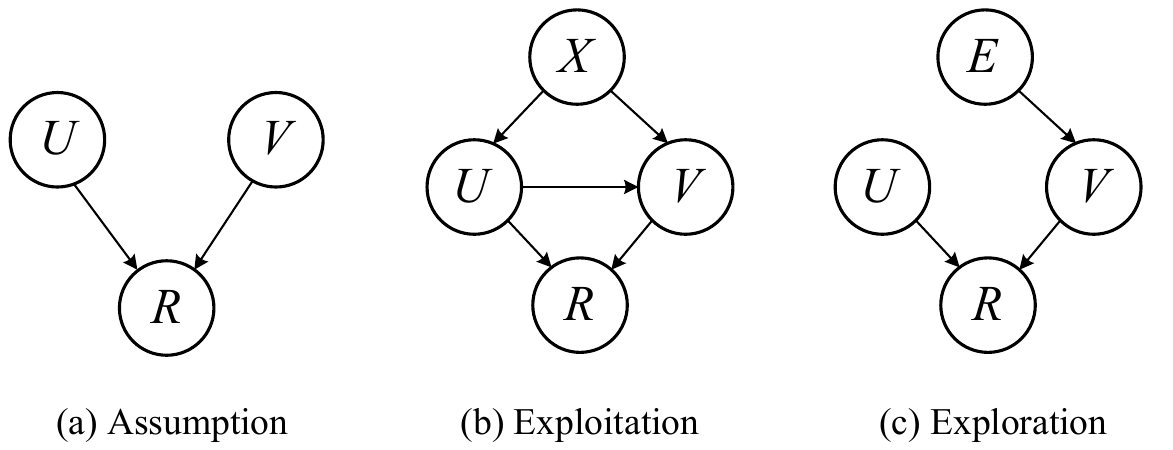}
\caption{$U$: user, $V$: item, $R$: rating, $X$: interaction history, $E$: a stochastic exploration strategy. (a) The hypothetical causal graph that traditional non-causality-based recommender systems assume to generate the collected rating data, where the exposure of item $V$ is independent of user $U$. (b) The true causal graph that generates the data collected by an exploitation-based strategy, where the exposure of item $V$ is dependent on the user's interaction history $X$. (c) The true causal graph that generates the data collected by an exploration-based strategy, where the exposure is based solely on the random exploration strategy $E$.}
\label{fig:exp}
\end{figure}

\noindent 
Although we establish the Deep-Deconf with Rubin's causal model in the main paper, we temporarily switch to Pearl's causal structural models to demonstrate how exploration eliminates confounding bias from the data collection perspective for better illustrative effects. Specifically, we provide three causal graphs as Fig. (\ref{fig:exp}).  Fig. (\ref{fig:exp}) - (a) is the hypothetical causal graph that traditional non-causality-based recommender systems assume to generate the collected rating data, where the exposure of item $V$ is independent of user $U$. Therefore, it can also be viewed as assuming the data are generated from randomized experiments from Rubin's causal perspective. Fig. (\ref{fig:exp}) - (b) shows a simple case of the true data generation process where the recommender systems used to collect the data greedily exploit the historical user ratings and user's preference to make recommendations. Note that If we use a naive model that assumes the data generation process of (a) to fit on rating data actually generated according to (b), the influence of unobserved confounders is mistakenly captured as the user preference, which leads to confounding bias in these models. Fig. (\ref{fig:exp}) - (c) shows the causal graph where the recommender system uses a random exploration strategy (i.e., randomly picking up an item to show the users) to collect data. In (c), we can find that since the item exposure is due solely to the random exploration, the exposure bias ($U \rightarrow V$) and confounding bias ($U \leftarrow X \rightarrow V$) is negated by the randomness. Therefore, models with the assumption of (a) can still be unbiased when trained on data collected by (c). Most recommender systems are a balance between exploitation in (b) and exploration in (c). Through comparisons among the three causal graphs, we analyze the exploration and exploitation, two commonly used strategies in reinforcement learning-based recommender systems, from a causal perspective; based on this, we provide a new justification for exploration strategies other than reducing the uncertainty in user preference estimations.

\definecolor{link}{HTML}{2D2F92}

\end{document}